\documentclass[11pt]{article}

\usepackage[utf8]{inputenc}%
\usepackage[T1]{fontenc}%
\usepackage[margin=1in]{geometry}%
\usepackage{graphicx}%
\usepackage{multirow}%
\usepackage{amsmath,amssymb,amsfonts}%
\usepackage{xcolor}%
\usepackage{textcomp}%
\usepackage{booktabs}%
\usepackage[numbers,sort&compress]{natbib}%
\usepackage[font=small,labelfont=bf,labelsep=period]{caption}%
\usepackage{etoolbox}%
\usepackage{array}%
\usepackage{tabularx}%
\usepackage{ragged2e}%
\usepackage{indentfirst}
\usepackage{placeins}
\usepackage{pgfplots}%
\pgfplotsset{compat=1.18}%
\usetikzlibrary{positioning,arrows.meta,decorations.pathreplacing}%
\usepgfplotslibrary{statistics}%
\usepackage[colorlinks=true,linkcolor=blue!60!black,citecolor=blue!60!black,urlcolor=blue!60!black]{hyperref}%
\usepackage{orcidlink}

\newcommand{\botrule}{\bottomrule}
\newcommand{\bmhead}[1]{\paragraph*{#1}}
\newcommand{\backmatter}{}
\providecommand{\bibcommenthead}{}
\newcommand{\tablenotes}{%
  \renewcommand{\footnotetext}[2][]{\par\smallskip{\footnotesize\noindent\ifstrempty{##1}{}{\textsuperscript{##1}}##2\par}}}
\AtBeginEnvironment{table}{\tablenotes\centering}

\hypersetup{bookmarksdepth=2}

\definecolor{pocA}{RGB}{0,114,178}
\definecolor{pocB}{RGB}{230,159,0}
\definecolor{pocC}{RGB}{0,158,115}
\newcolumntype{Y}{>{\RaggedRight\arraybackslash}X}
\newcolumntype{P}[1]{>{\RaggedRight\arraybackslash}p{#1}}
\newcommand{\medci}[3]{\begin{tabular}[t]{@{}c@{}}#1\\[-1pt]{\footnotesize[#2, #3]}\end{tabular}}
\AtBeginEnvironment{table}{\hyphenpenalty=10000\exhyphenpenalty=10000}

\newsavebox{\figbox}
\newenvironment{fitwidth}{\begin{lrbox}{\figbox}}{\end{lrbox}%
  \ifdim\wd\figbox>\linewidth\resizebox{\linewidth}{!}{\usebox{\figbox}}\else\usebox{\figbox}\fi}

\newcommand{\RAnalizadasA}{77}
\newcommand{\RAutCompSinHerrA}{16}
\newcommand{\RAutCompSinHerrB}{20}
\newcommand{\RAutCompSinHerrC}{61}
\newcommand{\RAutLlamadasC}{77}
\newcommand{\RAutSobreMaxMontoC}{24{,}000}
\newcommand{\RCajaCorrA}[1]{\addplot+[#1, boxplot prepared={lower whisker=1252.0, lower quartile=1650.8, median=2414.3, upper quartile=2666.0, upper whisker=4145.5}] coordinates {(0,4230.3) (0,4308.9) (0,4357.0) (0,4500.2) (0,4944.8) (0,4968.3) (0,4969.4) (0,4973.8) (0,5021.1) (0,5068.5) (0,5083.0) (0,5087.7) (0,5089.1) (0,5092.9) (0,5136.0) (0,5184.2) (0,5196.9) (0,5312.3) (0,5357.8) (0,5358.6) (0,5399.0) (0,5449.8) (0,5505.0) (0,5646.7) (0,6151.4) (0,6170.0) (0,6340.8) (0,6356.4) (0,7071.0) (0,8109.2)};}
\newcommand{\RCajaCorrB}[1]{\addplot+[#1, boxplot prepared={lower whisker=886.5, lower quartile=1661.4, median=2439.6, upper quartile=2774.9, upper whisker=4227.8}] coordinates {(0,4454.8) (0,4480.4) (0,4505.3) (0,4540.3) (0,4540.6) (0,4578.6) (0,4588.6) (0,4647.1) (0,4657.2) (0,4683.5) (0,4685.0) (0,4890.4) (0,5012.1) (0,5638.9)};}
\newcommand{\RCajaObsA}[1]{\addplot+[#1, boxplot prepared={lower whisker=1127.0, lower quartile=1842.2, median=2616.0, upper quartile=2916.2, upper whisker=4372.0}] coordinates {(0,4547.0) (0,4552.0) (0,4568.0) (0,4830.0) (0,5082.0) (0,5192.0) (0,5197.0) (0,5200.0) (0,5202.0) (0,5271.0) (0,5278.0) (0,5314.0) (0,5315.0) (0,5341.0) (0,5365.0) (0,5409.0) (0,5454.0) (0,5471.0) (0,5477.0) (0,5493.0) (0,5554.0) (0,5562.0) (0,5649.0) (0,5801.0) (0,5902.0) (0,6348.0) (0,6431.0) (0,6450.0) (0,7270.0) (0,8285.0)};}
\newcommand{\RCajaObsB}[1]{\addplot+[#1, boxplot prepared={lower whisker=1569.0, lower quartile=2222.8, median=3090.0, upper quartile=3508.5, upper whisker=5388.0}] coordinates {(0,5521.0) (0,5781.0) (0,5810.0) (0,5820.0) (0,5840.0) (0,6320.0)};}
\newcommand{\RCajaObsC}[1]{\addplot+[#1, boxplot prepared={lower whisker=181.4, lower quartile=1856.9, median=2444.8, upper quartile=3239.1, upper whisker=5283.6}] coordinates {(0,5314.7) (0,5485.7) (0,5488.7) (0,5724.8) (0,5760.6) (0,5765.9) (0,5831.4) (0,5860.7) (0,5923.6) (0,6045.9) (0,6285.9) (0,6737.6) (0,7096.4)};}
\newcommand{\RCensoTokensMed}{2{,}811}
\newcommand{\RCerA}{3.99}
\newcommand{\RCerB}{3.07}
\newcommand{\RCerC}{0.66}
\newcommand{\RCerEfectoDosTres}{0.72}
\newcommand{\RCerEfectoUnoDos}{0.39}
\newcommand{\RCerEfectoUnoTres}{0.76}
\newcommand{\RCerHiA}{4.42}
\newcommand{\RCerHiB}{3.40}
\newcommand{\RCerHiC}{0.91}
\newcommand{\RCerKwEpsilon}{0.51}
\newcommand{\RCerKwP}{$<$0.001}
\newcommand{\RCerLoA}{3.55}
\newcommand{\RCerLoB}{2.74}
\newcommand{\RCerLoC}{0.43}
\newcommand{\RCerPruebasA}{77}
\newcommand{\RCerWPBonfDosTres}{$<$0.001}
\newcommand{\RCerWPBonfUnoDos}{0.002}
\newcommand{\RCerWPBonfUnoTres}{$<$0.001}
\newcommand{\RColaFilaDifUnoTres}{160.4}
\newcommand{\RColaFilaDifVeredictoUnoTres}{no se distingue}
\newcommand{\RColaFilaMedC}{2{,}455.6}
\newcommand{\RCompCorregidoA}{1}
\newcommand{\RCompCorregidoB}{0}
\newcommand{\RCompCorregidoC}{7}
\newcommand{\RCompNA}{13}
\newcommand{\RCompNB}{13}
\newcommand{\RCompNC}{13}
\newcommand{\RCompRetractadoA}{7}
\newcommand{\RCompRetractadoB}{4}
\newcommand{\RCompRetractadoC}{0}
\newcommand{\RCompSttA}{2}
\newcommand{\RCompSttB}{4}
\newcommand{\RCompSttC}{13}
\newcommand{\RCompVisibleA}{8}
\newcommand{\RCompVisibleB}{4}
\newcommand{\RCompVisibleC}{7}
\newcommand{\RCoordCapasA}{(1003.5,3) (2.6,1) (185.8,7) (559.0,6) (231.0,5) (687.0,4) (297.0,2)}
\newcommand{\RCoordCapasB}{(1121.0,3) (213.8,1) (423.1,7) (577.5,6) (236.6,5) (680.3,4) (219.8,2)}
\newcommand{\RCoordCapasC}{(923.8,3) (71.0,1)}
\newcommand{\RCorrHiA}{2{,}437.7}
\newcommand{\RCorrHiB}{2{,}509.7}
\newcommand{\RCorrLoA}{2{,}378.9}
\newcommand{\RCorrLoB}{2{,}373.8}
\newcommand{\RCorrMedA}{2{,}414.3}
\newcommand{\RCorrMedB}{2{,}439.6}
\newcommand{\RCorridasA}{90}
\newcommand{\RDifAbsDosTres}{645.2}
\newcommand{\RDifAbsUnoDos}{474.0}
\newcommand{\RDifAbsUnoTres}{171.2}
\newcommand{\RDifCorrHiUnoDos}{43.1}
\newcommand{\RDifCorrLoUnoDos}{$-$106.6}
\newcommand{\RDifCorrUnoDos}{$-$25.2}
\newcommand{\RDifCorrVeredictoUnoDos}{equivalentes en la práctica}
\newcommand{\RDifDosTres}{645.2}
\newcommand{\RDifHastaResHiUnoTres}{279.0}
\newcommand{\RDifHastaResLoUnoTres}{$-$156.6}
\newcommand{\RDifHastaResUnoTres}{83.0}
\newcommand{\RDifHastaResVeredictoUnoTres}{no se distingue}
\newcommand{\RDifHiDosTres}{825.3}
\newcommand{\RDifHiUnoDos}{$-$404.5}
\newcommand{\RDifHiUnoTres}{322.5}
\newcommand{\RDifLoDosTres}{475.1}
\newcommand{\RDifLoUnoDos}{$-$583.0}
\newcommand{\RDifLoUnoTres}{$-$3.2}
\newcommand{\RDifSinSalidaHiUnoTres}{396.9}
\newcommand{\RDifSinSalidaLoUnoTres}{74.1}
\newcommand{\RDifSinSalidaUnoTres}{239.5}
\newcommand{\RDifSinSalidaVeredictoUnoTres}{distinta de cero; no se sabe si pasa de 200 ms}
\newcommand{\RDifUnoDos}{$-$474.0}
\newcommand{\RDifUnoTres}{171.2}
\newcommand{\RDifVeredictoDosTres}{distinta de cero y mayor que 200 ms}
\newcommand{\RDifVeredictoUnoDos}{distinta de cero y mayor que 200 ms}
\newcommand{\RDifVeredictoUnoTres}{no se distingue}
\newcommand{\RDispIqrA}{1{,}074}
\newcommand{\RDispIqrSinHerrA}{310}
\newcommand{\REncimaA}{1}
\newcommand{\REncimaC}{26}
\newcommand{\REntCpu}{AMD Ryzen 5 3400G with Radeon Vega Graphics}
\newcommand{\REntNucleos}{8}
\newcommand{\REntSistema}{Windows 11 (10.0.26200)}
\newcommand{\REtapaLlmA}{687}
\newcommand{\REtapaLlmB}{680}
\newcommand{\REtapaSttA}{231}
\newcommand{\REtapaSttB}{237}
\newcommand{\RFalsosA}{513}
\newcommand{\RFalsosB}{43}
\newcommand{\RFuncCompensaA}{122}
\newcommand{\RFuncCompensaB}{123}
\newcommand{\RFuncCompensaC}{31}
\newcommand{\RGuionAlt}{7}
\newcommand{\RGuionBase}{40}
\newcommand{\RGuionLineas}{47}
\newcommand{\RGuionTurnosConRamas}{6}
\newcommand{\RHastaResMedA}{2{,}623.5}
\newcommand{\RHastaResMedB}{3{,}095.0}
\newcommand{\RHastaResMedC}{2{,}540.5}
\newcommand{\RHerrFrac}{65}
\newcommand{\RHerrHablo}{79}
\newcommand{\RHerrMs}{653}
\newcommand{\RHerrN}{121}
\newcommand{\RIdaA}{185.8}
\newcommand{\RIdaB}{423.1}
\newcommand{\RIntAcuApoyoA}{3}
\newcommand{\RIntAcuApoyoB}{2}
\newcommand{\RIntAcuApoyoC}{0}
\newcommand{\RIntAcuNoAceptadosA}{5}
\newcommand{\RIntAcuNoAceptadosB}{7}
\newcommand{\RIntAcuNoAceptadosC}{1}
\newcommand{\RIntAcuRegistradosA}{12}
\newcommand{\RIntAcuRegistradosB}{13}
\newcommand{\RIntAcuRegistradosC}{10}
\newcommand{\RIntApoFracA}{26/28}
\newcommand{\RIntApoFracB}{21/22}
\newcommand{\RIntApoFracC}{20/29}
\newcommand{\RIntApoPauMedA}{2{,}149}
\newcommand{\RIntApoPauMedB}{3{,}331}
\newcommand{\RIntApoPauMedC}{1{,}241}
\newcommand{\RIntApoPauQaA}{1{,}534}
\newcommand{\RIntApoPauQaB}{2{,}458}
\newcommand{\RIntApoPauQaC}{1{,}008}
\newcommand{\RIntApoPauQbA}{2{,}923}
\newcommand{\RIntApoPauQbB}{4{,}142}
\newcommand{\RIntApoPauQbC}{2{,}038}
\newcommand{\RIntApoRetomaA}{26/26}
\newcommand{\RIntApoRetomaB}{21/21}
\newcommand{\RIntApoRetomaC}{19/20}
\newcommand{\RIntCtlFracA}{0/28}
\newcommand{\RIntCtlFracB}{0/22}
\newcommand{\RIntCtlFracC}{13/29}
\newcommand{\RIntCtlPauMedC}{1{,}308}
\newcommand{\RIntDetMedA}{344}
\newcommand{\RIntDetMedB}{878}
\newcommand{\RIntDetMedC}{1{,}242}
\newcommand{\RIntDetQaA}{234}
\newcommand{\RIntDetQaB}{856}
\newcommand{\RIntDetQaC}{391}
\newcommand{\RIntDetQbA}{448}
\newcommand{\RIntDetQbB}{903}
\newcommand{\RIntDetQbC}{1{,}387}
\newcommand{\RIntIntFracA}{26/27}
\newcommand{\RIntIntFracB}{22/22}
\newcommand{\RIntIntFracC}{25/26}
\newcommand{\RIntLlamadasA}{15}
\newcommand{\RIntLlamadasB}{15}
\newcommand{\RIntLlamadasC}{15}
\newcommand{\RIntPruebaAlfa}{0.00625}
\newcommand{\RIntPruebaApoyoControlTresP}{0.130}
\newcommand{\RIntPruebaPausaTresDosP}{$<$0.001}
\newcommand{\RIntPruebaSigueTresDosP}{0.090}
\newcommand{\RIntPruebaSigueTresUnoP}{0.031}
\newcommand{\RIntRetCA}{5}
\newcommand{\RIntRetCB}{5}
\newcommand{\RIntRetCC}{10}
\newcommand{\RIntRetNA}{6}
\newcommand{\RIntRetNB}{7}
\newcommand{\RIntRetNC}{2}
\newcommand{\RIntRetRA}{5}
\newcommand{\RIntRetRB}{3}
\newcommand{\RIntRetRC}{4}
\newcommand{\RIntRetTA}{10}
\newcommand{\RIntRetTB}{5}
\newcommand{\RIntRetTC}{1}
\newcommand{\RIntRetTotalA}{26}
\newcommand{\RIntRetTotalB}{21}
\newcommand{\RIntRetTotalC}{19}
\newcommand{\RKwN}{231}
\newcommand{\RLineaDifDosTres}{497}
\newcommand{\RLineaDifHiDosTres}{600}
\newcommand{\RLineaDifHiUnoTres}{127}
\newcommand{\RLineaDifLoDosTres}{323}
\newcommand{\RLineaDifLoUnoTres}{$-$172}
\newcommand{\RLineaDifUnoTres}{$-$19}
\newcommand{\RLineaDifVeredictoDosTres}{distinta de cero y mayor que 200 ms}
\newcommand{\RLineaDifVeredictoUnoTres}{equivalentes en la práctica}
\newcommand{\RLineaMasRapidaUnoTres}{23}
\newcommand{\RLineaN}{38}
\newcommand{\RLlamadasA}{77}
\newcommand{\RLlamadasB}{77}
\newcommand{\RLlamadasC}{77}
\newcommand{\RLlamadasEncimaC}{20}
\newcommand{\RLtAntesFin}{11.16}
\newcommand{\RLtAntesIni}{8.85}
\newcommand{\RLtApoyoFin}{10.70}
\newcommand{\RLtApoyoIni}{10.37}
\newcommand{\RLtCallaSeg}{0.8}
\newcommand{\RLtCuelgaIni}{25.23}
\newcommand{\RLtDespuesFin}{22.99}
\newcommand{\RLtDespuesIni}{13.41}
\newcommand{\RLtEje}{30}
\newcommand{\RLtLineaFin}{5.45}
\newcommand{\RLtLineaIni}{0.00}
\newcommand{\RLtMonto}{16{,}500}
\newcommand{\RLtRegistroIni}{12.62}
\newcommand{\RLtRegistroSeg}{2.3}
\newcommand{\RLtSigFin}{27.18}
\newcommand{\RLtSigIni}{25.02}
\newcommand{\RMezclaDifHiUnoTres}{295.5}
\newcommand{\RMezclaDifLoUnoTres}{52.1}
\newcommand{\RMezclaDifUnoTres}{171.2}
\newcommand{\RMezclaDifVeredictoUnoTres}{distinta de cero; no se sabe si pasa de 200 ms}
\newcommand{\RMontosA}{13.01}
\newcommand{\RMontosB}{12.61}
\newcommand{\RMontosC}{2.86}
\newcommand{\RMontosEfectoDosTres}{0.29}
\newcommand{\RMontosEfectoUnoDos}{0.04}
\newcommand{\RMontosEfectoUnoTres}{0.32}
\newcommand{\RMontosHiA}{17.69}
\newcommand{\RMontosHiB}{17.54}
\newcommand{\RMontosHiC}{6.63}
\newcommand{\RMontosKwEpsilon}{0.07}
\newcommand{\RMontosKwP}{0.007}
\newcommand{\RMontosLoA}{8.26}
\newcommand{\RMontosLoB}{7.55}
\newcommand{\RMontosLoC}{0.00}
\newcommand{\RMontosNA}{123}
\newcommand{\RMontosNB}{111}
\newcommand{\RMontosNC}{140}
\newcommand{\RMontosPruebasA}{45}
\newcommand{\RMontosWPBonfDosTres}{0.018}
\newcommand{\RMontosWPBonfDosTresTxt}{$p=0.018$}
\newcommand{\RMontosWPBonfUnoDos}{1.000}
\newcommand{\RMontosWPBonfUnoDosTxt}{$p=1.000$}
\newcommand{\RMontosWPBonfUnoTres}{0.006}
\newcommand{\RMontosWPBonfUnoTresTxt}{$p=0.006$}
\newcommand{\RNA}{558}
\newcommand{\RNB}{536}
\newcommand{\RNC}{563}
\newcommand{\RObsHiA}{2{,}663.0}
\newcommand{\RObsHiB}{3{,}200.0}
\newcommand{\RObsHiC}{2{,}608.2}
\newcommand{\RObsLoA}{2{,}582.0}
\newcommand{\RObsLoB}{3{,}038.0}
\newcommand{\RObsLoC}{2{,}303.1}
\newcommand{\RObsMedA}{2{,}616.0}
\newcommand{\RObsMedB}{3{,}090.0}
\newcommand{\RObsMedC}{2{,}444.8}
\newcommand{\RObsNoventaycincoA}{4{,}554.4}
\newcommand{\RObsNoventaycincoB}{4{,}630.2}
\newcommand{\RObsNoventaycincoC}{4{,}813.8}
\newcommand{\RPares}{77}
\newcommand{\RPrimerDifHiUnoTres}{559.0}
\newcommand{\RPrimerDifLoUnoTres}{348.1}
\newcommand{\RPrimerDifUnoTresIc}{471.4}
\newcommand{\RPrimerDifVeredictoUnoTres}{distinta de cero y mayor que 200 ms}
\newcommand{\RPrimerFracA}{12.4}
\newcommand{\RPrimerFracB}{4.7}
\newcommand{\RPrimerFracC}{25.0}
\newcommand{\RPrimerHiA}{2{,}631.1}
\newcommand{\RPrimerHiB}{3{,}162.2}
\newcommand{\RPrimerHiC}{2{,}223.9}
\newcommand{\RPrimerLoA}{2{,}537.2}
\newcommand{\RPrimerLoB}{3{,}002.9}
\newcommand{\RPrimerLoC}{2{,}032.3}
\newcommand{\RPrimerMedA}{2{,}597.2}
\newcommand{\RPrimerMedB}{3{,}070.9}
\newcommand{\RPrimerMedC}{2{,}125.8}
\newcommand{\RPropiaA}{2{,}371.0}
\newcommand{\RPropiaB}{2{,}417.3}
\newcommand{\RPropiaFaltaA}{245}
\newcommand{\RPropiaFaltaB}{673}
\newcommand{\RPropiaFaltaPctA}{9}
\newcommand{\RPropiaFaltaPctB}{22}
\newcommand{\RReconMedA}{2{,}620.2}
\newcommand{\RReconMedB}{3{,}092.1}
\newcommand{\RReconObsMedA}{2{,}616.0}
\newcommand{\RReconObsMedB}{3{,}090.0}
\newcommand{\RRecuperadosC}{49}
\newcommand{\RRecuperadosPctC}{9}
\newcommand{\RRedMed}{42}
\newcommand{\RRelojDerivaLapsoMed}{119}
\newcommand{\RRelojDerivaMed}{811}
\newcommand{\RRelojEjEntrega}{0.167}
\newcommand{\RRelojEjEntregaMs}{167}
\newcommand{\RRelojEjModelo}{0.800}
\newcommand{\RRelojEjModeloMs}{800}
\newcommand{\RRelojEjSalida}{0.309}
\newcommand{\RRelojEjSalidaMs}{309}
\newcommand{\RRelojEjTotalMs}{2{,}383}
\newcommand{\RRelojEjUltima}{1.107}
\newcommand{\RRelojEjUltimaMs}{1{,}107}
\newcommand{\RRelojEntregaConHerrMed}{185}
\newcommand{\RRelojEntregaMed}{171}
\newcommand{\RRelojEntregaQa}{115}
\newcommand{\RRelojEntregaQb}{209}
\newcommand{\RRelojEntregaSinHerrMed}{169}
\newcommand{\RRelojLatConHerrMed}{3{,}068}
\newcommand{\RRelojLatMed}{2{,}448}
\newcommand{\RRelojLatQa}{1{,}866}
\newcommand{\RRelojLatQb}{3{,}234}
\newcommand{\RRelojLatSinHerrMed}{2{,}147}
\newcommand{\RRelojModeloConHerrMed}{1{,}400}
\newcommand{\RRelojModeloMed}{800}
\newcommand{\RRelojModeloQa}{400}
\newcommand{\RRelojModeloQb}{1{,}600}
\newcommand{\RRelojModeloSinHerrMed}{600}
\newcommand{\RRelojModeloValor}{800.0}
\newcommand{\RRelojN}{560}
\newcommand{\RRelojNConHerr}{184}
\newcommand{\RRelojNSinHerr}{376}
\newcommand{\RRelojRestoConHerrMed}{1{,}563}
\newcommand{\RRelojRestoMed}{1{,}537}
\newcommand{\RRelojRestoQa}{1{,}424}
\newcommand{\RRelojRestoQb}{1{,}656}
\newcommand{\RRelojRestoSinHerrMed}{1{,}526}
\newcommand{\RRelojRestoValor}{1537.3}
\newcommand{\RRelojSalidaConHerrMed}{309}
\newcommand{\RRelojSalidaMed}{297}
\newcommand{\RRelojSalidaQa}{233}
\newcommand{\RRelojSalidaQb}{344}
\newcommand{\RRelojSalidaSinHerrMed}{291}
\newcommand{\RRelojServicioConHerrMed}{1{,}380}
\newcommand{\RRelojServicioMed}{1{,}367}
\newcommand{\RRelojServicioQa}{1{,}284}
\newcommand{\RRelojServicioQb}{1{,}464}
\newcommand{\RRelojServicioSinHerrMed}{1{,}358}
\newcommand{\RRelojSumaMedMs}{2{,}350}
\newcommand{\RRelojUltimaConHerrMed}{1{,}078}
\newcommand{\RRelojUltimaMed}{1{,}082}
\newcommand{\RRelojUltimaQa}{990}
\newcommand{\RRelojUltimaQb}{1{,}178}
\newcommand{\RRelojUltimaSinHerrMed}{1{,}085}
\newcommand{\RReplicas}{10{,}000}
\newcommand{\RRojasA}{2}
\newcommand{\RRojasB}{5}
\newcommand{\RRojasC}{8}
\newcommand{\RRtConA}{2{,}810.5}
\newcommand{\RRtConB}{3{,}231.0}
\newcommand{\RRtConC}{3{,}068.9}
\newcommand{\RRtDeltaA}{1{,}003.5}
\newcommand{\RRtDeltaB}{1{,}121.0}
\newcommand{\RRtDeltaC}{923.8}
\newcommand{\RRtDeltaHiA}{1{,}062.5}
\newcommand{\RRtDeltaHiB}{1{,}225.0}
\newcommand{\RRtDeltaHiC}{1{,}185.9}
\newcommand{\RRtDeltaLoA}{953.0}
\newcommand{\RRtDeltaLoB}{980.0}
\newcommand{\RRtDeltaLoC}{779.7}
\newcommand{\RRtDeltaMax}{1.1}
\newcommand{\RRtDeltaMin}{0.9}
\newcommand{\RRtFracA}{65.2}
\newcommand{\RRtFracB}{65.1}
\newcommand{\RRtFracC}{32.9}
\newcommand{\RRtFracPrevA}{58.4}
\newcommand{\RRtFracPrevB}{59.1}
\newcommand{\RRtFracPrevC}{32.5}
\newcommand{\RRtNConA}{364}
\newcommand{\RRtNConB}{349}
\newcommand{\RRtNConC}{185}
\newcommand{\RRtSinA}{1{,}807.0}
\newcommand{\RRtSinB}{2{,}110.0}
\newcommand{\RRtSinC}{2{,}145.1}
\newcommand{\RRtSinHiA}{1{,}839.0}
\newcommand{\RRtSinHiB}{2{,}226.0}
\newcommand{\RRtSinHiC}{2{,}276.5}
\newcommand{\RRtSinLoA}{1{,}760.0}
\newcommand{\RRtSinLoB}{2{,}039.0}
\newcommand{\RRtSinLoC}{1{,}998.7}
\newcommand{\RSensPBonfSm}{0.003}
\newcommand{\RSolapesRetomoC}{22}
\newcommand{\RTramoEjFinA}{0.559}
\newcommand{\RTramoEjFinB}{0.578}
\newcommand{\RTramoEjFinMsA}{559}
\newcommand{\RTramoEjFinMsB}{578}
\newcommand{\RTramoEjIdaA}{0.165}
\newcommand{\RTramoEjIdaB}{0.407}
\newcommand{\RTramoEjIdaMsA}{165}
\newcommand{\RTramoEjIdaMsB}{407}
\newcommand{\RTramoEjLlmA}{0.661}
\newcommand{\RTramoEjLlmB}{0.491}
\newcommand{\RTramoEjLlmMsA}{661}
\newcommand{\RTramoEjLlmMsB}{491}
\newcommand{\RTramoEjPropiaMsA}{1{,}626}
\newcommand{\RTramoEjPropiaMsB}{1{,}294}
\newcommand{\RTramoEjSttA}{0.216}
\newcommand{\RTramoEjSttB}{0.229}
\newcommand{\RTramoEjSttMsA}{216}
\newcommand{\RTramoEjSttMsB}{229}
\newcommand{\RTramoEjTotalMsA}{1{,}793}
\newcommand{\RTramoEjTotalMsB}{1{,}911}
\newcommand{\RTramoEjTtsA}{0.406}
\newcommand{\RTramoEjTtsB}{0.225}
\newcommand{\RTramoEjTtsMsA}{406}
\newcommand{\RTramoEjTtsMsB}{225}
\newcommand{\RTramoEjVueltaA}{0.003}
\newcommand{\RTramoEjVueltaB}{0.210}
\newcommand{\RTramoEjVueltaMsA}{2}
\newcommand{\RTramoEjVueltaMsB}{210}
\newcommand{\RTramoFinCortoMedB}{578}
\newcommand{\RTramoFinLargoMedB}{2{,}501}
\newcommand{\RTramoFinLargoPctB}{31.2}
\newcommand{\RTramoSttAntesPctA}{99.3}
\newcommand{\RTramoSttAntesPctB}{100.0}
\newcommand{\RTramoTraslapeMedB}{189}
\newcommand{\RTransporteA}{188}
\newcommand{\RTransporteB}{637}
\newcommand{\RValBDosMilMediana}{229.4}
\newcommand{\RValBDosMilMedidos}{38}
\newcommand{\RValBDosMilTurnos}{40}
\newcommand{\RValBIdaNormal}{137.0}
\newcommand{\RValBMilMediana}{240.9}
\newcommand{\RValBMilMedidos}{39}
\newcommand{\RValBMilTurnos}{40}
\newcommand{\RValBSeisMediana}{268.9}
\newcommand{\RValBSeisMedidos}{39}
\newcommand{\RValBSeisTurnos}{40}
\newcommand{\RValBSuaveMediana}{287.9}
\newcommand{\RValBSuaveMedidos}{39}
\newcommand{\RValBSuaveTurnos}{40}
\newcommand{\RValBTreintaMediana}{259.9}
\newcommand{\RValBTreintaMedidos}{39}
\newcommand{\RValBTreintaTurnos}{40}
\newcommand{\RValBTresMediana}{320.1}
\newcommand{\RValBTresMedidos}{39}
\newcommand{\RValBTresTurnos}{40}
\newcommand{\RValBVueltaNormal}{117.7}
\newcommand{\RValCDosMilTurnos}{40}
\newcommand{\RValCMilTurnos}{40}
\newcommand{\RValCSeisTurnos}{40}
\newcommand{\RValCSuaveTurnos}{40}
\newcommand{\RValCTreintaTurnos}{40}
\newcommand{\RValCTresTurnos}{40}
\newcommand{\RValCorridas}{30}
\newcommand{\RValDDosMilMediana}{$-$46.4}
\newcommand{\RValDDosMilMedidos}{40}
\newcommand{\RValDMilMediana}{$-$42.9}
\newcommand{\RValDMilMedidos}{40}
\newcommand{\RValDSeisMediana}{$-$2.8}
\newcommand{\RValDSeisMedidos}{40}
\newcommand{\RValDSuaveMediana}{2.2}
\newcommand{\RValDSuaveMedidos}{40}
\newcommand{\RValDTreintaMediana}{$-$11.9}
\newcommand{\RValDTreintaMedidos}{40}
\newcommand{\RValDTresMediana}{$-$40.9}
\newcommand{\RValDTresMedidos}{40}
\newcommand{\RValDosMilMediana}{$-$48.0}
\newcommand{\RValDosMilMedidos}{40}
\newcommand{\RValDosMilTurnos}{40}
\newcommand{\RValErrMaxAbs}{86.4}
\newcommand{\RValMilMediana}{$-$42.3}
\newcommand{\RValMilMedidos}{40}
\newcommand{\RValMilTurnos}{40}
\newcommand{\RValSeisMediana}{$-$10.1}
\newcommand{\RValSeisMedidos}{40}
\newcommand{\RValSeisTurnos}{40}
\newcommand{\RValSuaveMediana}{$-$2.3}
\newcommand{\RValSuaveMedidos}{40}
\newcommand{\RValSuaveTurnos}{40}
\newcommand{\RValTreintaMediana}{$-$8.5}
\newcommand{\RValTreintaMedidos}{40}
\newcommand{\RValTreintaTurnos}{40}
\newcommand{\RValTresMediana}{$-$46.0}
\newcommand{\RValTresMedidos}{38}
\newcommand{\RValTresTurnos}{40}
\newcommand{\RValTurnos}{240}
\newcommand{\RViaLineas}{4}
\newcommand{\RViaRtcFueraMed}{1{,}368}
\newcommand{\RViaRtcMed}{1{,}342}
\newcommand{\RViaRtcModeloMed}{0}
\newcommand{\RViaSesiones}{3}
\newcommand{\RViaWsFueraMed}{1{,}394}
\newcommand{\RViaWsMed}{1{,}228}
\newcommand{\RViaWsModeloMed}{$-$200}
\newcommand{\RViaWsN}{12}
\newcommand{\RVueltaA}{2.6}
\newcommand{\RVueltaB}{213.8}
\newcommand{\RVueltaC}{71.0}
\newcommand{\RWPBonfUnoTres}{1.000}
\newcommand{\RWerA}{6.03}
\newcommand{\RWerB}{4.93}
\newcommand{\RWerC}{1.47}
\newcommand{\RWerEfectoDosTres}{0.69}
\newcommand{\RWerEfectoUnoDos}{0.40}
\newcommand{\RWerEfectoUnoTres}{0.73}
\newcommand{\RWerFueraA}{0}
\newcommand{\RWerFueraB}{4}
\newcommand{\RWerFueraC}{13}
\newcommand{\RWerHiA}{6.63}
\newcommand{\RWerHiB}{5.36}
\newcommand{\RWerHiC}{1.89}
\newcommand{\RWerKwEpsilon}{0.46}
\newcommand{\RWerKwP}{$<$0.001}
\newcommand{\RWerKwPTxt}{$p<0.001$}
\newcommand{\RWerLoA}{5.42}
\newcommand{\RWerLoB}{4.49}
\newcommand{\RWerLoC}{1.09}
\newcommand{\RWerNA}{5{,}905}
\newcommand{\RWerNB}{5{,}725}
\newcommand{\RWerNC}{5{,}913}
\newcommand{\RWerPruebasA}{77}
\newcommand{\RWerVecesA}{4.1}
\newcommand{\RWerVecesB}{3.3}
\newcommand{\RWerWPBonfDosTres}{$<$0.001}
\newcommand{\RWerWPBonfDosTresTxt}{$p<0.001$}
\newcommand{\RWerWPBonfUnoDos}{0.001}
\newcommand{\RWerWPBonfUnoDosTxt}{$p=0.001$}
\newcommand{\RWerWPBonfUnoTres}{$<$0.001}
\newcommand{\RWerWPBonfUnoTresTxt}{$p<0.001$}

\makeatletter
\newcommand{\ver@eq}{equivalentes en la práctica}
\newcommand{\ver@nd}{no se distingue}
\newcommand{\ver@menor}{distinta de cero, menor que 200 ms}
\newcommand{\ver@mayor}{distinta de cero y mayor que 200 ms}
\newcommand{\ver@cruza}{distinta de cero; no se sabe si pasa de 200 ms}
\newcommand{\verdict}[1]{%
  \ifx#1\ver@eq equivalent in practice%
  \else\ifx#1\ver@nd not distinguishable from zero%
  \else\ifx#1\ver@menor different from zero, below 200~ms%
  \else\ifx#1\ver@mayor different from zero, above 200~ms%
  \else\ifx#1\ver@cruza different from zero; may or may not exceed 200~ms%
  \else\textbf{??}\PackageWarning{paper}{Unknown verdict}%
  \fi\fi\fi\fi\fi}
\makeatother

\begin{document}

\title{Latency, Speech Recognition and Turn-Taking in Voice Agents for Spanish Phone Negotiation: An Externally Measured Comparison of Pipecat, LiveKit and GPT-Live}

\author{%
Jovan Popoca Moran$^{1}$, Eduardo de Avila-Armenta$^{1,2}$\,\orcidlink{0000-0003-3461-7722}, Raúl Alejandro Aquino-Ibarra$^{1}$,\\
Jorge Octavio Nicolás-Díaz$^{1}$, Umberto Leon-Dominguez$^{1,3*}$\,\orcidlink{0000-0001-5170-2553}\\[4pt]
{\normalsize $^{1}$AI Lab, Freight Technologies Inc., Monterrey, N.L., Mexico}\\
{\normalsize $^{2}$Tecnológico de Monterrey, School of Engineering and Sciences, Monterrey, N.L., Mexico}\\
{\normalsize $^{3}$School of Psychology, University of Monterrey, Monterrey, N.L., Mexico}}

\date{\today}

\maketitle

{\renewcommand{\thefootnote}{*}\footnotetext{Corresponding author. Email: \texttt{umberto.leon@udem.edu}}}

\begin{abstract}
A voice agent that negotiates by phone must answer quickly, understand spoken amounts and cope with a caller who talks over it.
We compared three architectures for such an agent, negotiating freight rates in Spanish: two cascades (speech recognition, language model and speech synthesis) orchestrated by Pipecat and by LiveKit Agents, and GPT-Live, an OpenAI speech model that listens and speaks without separate stages.
A program played the carrier with the same \RGuionLineas{} recordings of one speaker for all agents, without the telephone network, in \RLlamadasA{} calls per architecture over five negotiation scenarios, and measured latency on the listener's side with a validated instrument.
Median latency to the content response was \RObsMedA~ms (Pipecat), \RObsMedB~ms (LiveKit) and \RObsMedC~ms (GPT-Live); the confidence interval of GPT-Live's \RDifAbsUnoTres~ms advantage over Pipecat includes zero.
According to its own timestamps, GPT-Live began to answer \RRelojModeloMed~ms after the end of the line, close to the 0.8~s published by OpenAI; most of the remaining wait, about \RRelojServicioMed~ms and nearly constant, is the audio's round trip through OpenAI's service.
A tool call added \RRtDeltaMin{}--\RRtDeltaMax~s in every architecture, more than any difference between them, and the frameworks' self-reported latency fell short of the listener's wait by \RPropiaFaltaPctA\,\% (Pipecat) and \RPropiaFaltaPctB\,\% (LiveKit).
Word error rates were low (\RWerA, \RWerB{} and \RWerC\,\%), but the error on amounts was \RMontosA\,\% and \RMontosB\,\% in the cascades, against \RMontosC\,\% in GPT-Live.
With default settings, Pipecat and LiveKit stopped at almost every backchannel and often took it as the carrier's answer: an ``ok'' spoken over the agent ended in a registered agreement in \RIntAcuApoyoA{} of \RIntLlamadasA{} Pipecat calls, \RIntAcuApoyoB{} of \RIntLlamadasB{} LiveKit calls and no GPT-Live call (descriptive counts, not tested).
Latency should therefore be measured on the listener's side, and a cascade needs safeguards for spoken amounts and backchannels.

\end{abstract}

\section{Introduction}\label{sec:intro}

Spoken dialogue agents built around large language models (LLMs) are usually assembled in one of two ways~\cite{arora2025survey}.
In a cascade, a speech recognizer (speech-to-text, STT) turns the caller's audio into text, an LLM decides the response and a synthesizer (text-to-speech, TTS) turns that response into audio.
Each stage can be chosen, measured and replaced separately, but their delays add up, and the system needs a separate detector to decide when the person has finished speaking.
In a native speech model, a single model takes audio in and produces audio out, and full-duplex models of this kind can listen while they speak~\cite{nguyen2023dgslm,zhang2023speechgpt,defossez2024moshi}; their internal stages, however, cannot be observed or tuned.
Open-source frameworks such as Pipecat~\cite{pipecat} and LiveKit Agents~\cite{livekitagents} orchestrate cascades, and providers now offer native speech models through their APIs~\cite{gptlive}, so a team that builds a voice agent has to choose between the two.

Speaking in real time raises problems that a text-based agent does not face.
In human conversation, the gap between one speaker's turn and the next is short: across ten typologically diverse languages, the most frequent gap lies between 0 and 200~ms~\cite{stivers2009}, which implies that listeners plan their reply while the other person is still speaking~\cite{levinson2015}.
Overlaps are also frequent~\cite{heldner2010}, and many of them are backchannels, short tokens such as ``uh-huh'' or ``ok'' with which the listener signals attention without claiming the floor~\cite{yngve1970,ward2000}.
A spoken agent must therefore decide when the caller has finished, when to yield to an interruption and when to keep talking through a backchannel, problems that a large body of work on turn-taking addresses~\cite{skantze2021,ekstedt2020turngpt,ekstedt2022vap}.
How long the agent takes to answer also shapes how users judge it, and short fillers can soften a longer wait~\cite{shiwa2009,maslych2025}.
Speech recognition adds a further difficulty: the word error rate (WER) is a weak indicator of how well a system understands what was said~\cite{wang2003wer,morris2004}, and errors on a few critical words, such as named entities, can matter far more than the overall rate suggests~\cite{benjannet2015}.

These requirements are concrete in the application that motivated this study.
SPOTS, a system for spot freight, already negotiates rates with carriers over WhatsApp, a task that belongs to the broader family of negotiation dialogues~\cite{lewis2017deal}.
Each negotiation starts from a configuration with a target price and a maximum price.
Transitioning this negotiation to a voice channel, however, introduces a significant architectural challenge.
If there is no agreement after 30 minutes on WhatsApp, a voice agent calls the carrier on behalf of the shipper, inheriting the same configuration to negotiate in Spanish.
This modality adds two critical requirements absent in text-based communication: the agent must answer in real time, before the silence becomes uncomfortable, and it must accurately process speech over a telephone line, which presents poorer audio quality than a standard recording.
To determine which architectural approach best addresses these constraints, an experiment was designed to compare three distinct voice architectures under identical conditions.
The first architecture is a cascade orchestrated by the open-source framework Pipecat~\cite{pipecat}.
The second architecture is an identical cascade, using the same underlying providers, orchestrated by LiveKit Agents~\cite{livekitagents}.
The third architecture employs GPT-Live~\cite{gptlive}, an OpenAI speech model that listens and speaks without separate STT or TTS stages, and which delegates tool calls to the same LLM used by the other two architectures.
Hereafter, each architecture is named after its framework or model: Pipecat, LiveKit and GPT-Live.

Comparisons between voice frameworks are usually based on demonstrations, on tests of a single component or on the latency that each framework reports about itself.
Benchmarks for full-duplex models evaluate turn-taking behaviours such as pause handling, backchannelling and interruptions~\cite{lin2025fdbench}, and the latency that OpenAI publishes for GPT-Live comes from one such benchmark~\cite{gptlive}.
In the present study, the three agents heard exactly the same audio: recorded carrier lines, played by a program instead of a person.
Latency was measured outside the frameworks, on the listener's side, with the same instrument for all three (a software measurement harness: a program that plays the carrier's lines and timestamps the end of each line and the start of the agent's response), and that instrument was checked against a simulated agent with a known delay over the audio path of each architecture.

The study measures what distinguishes one voice technology from another: response latency, speech recognition and turn-taking, including what each agent does when the carrier talks over it.
It does not measure whether the agent negotiates well.
The negotiation is the vehicle of the conversation: it brings domain vocabulary (amounts, places, trade terms) and forces the agent to call tools in the middle of its response.
The three agents use the same instructions and tools, but not the same conversational model: in Pipecat and LiveKit the conversation is handled by \texttt{gpt-4.1-mini}, and in GPT-Live by \texttt{gpt-live-1}, which hands tools over to \texttt{gpt-4.1-mini}.
Each architecture is compared as a whole, with its default configuration.

The study makes four contributions.
First, it provides an external, validated measurement protocol that can be repeated with other architectures.
Second, it compares latency while separating the weight of transport and of tool calls.
Third, it shows the gap between the WER and the error on amounts, which are the words that cost money in a negotiation.
Fourth, it describes what each architecture does, with its default configuration, when the carrier says ``ajá'' (``uh-huh'') while the agent is speaking.
The results support one central message: under identical, externally measured conditions, the three architectures differed less in latency than in how they handled spoken amounts and backchannels, and the latency that the frameworks reported about themselves hid most of the transport cost.
Section~\ref{sec:metodos} describes the materials and methods, Section~\ref{sec:resultados} presents the results, Section~\ref{sec:discusion} discusses them together with their limitations, and Section~\ref{sec:conclusion} concludes.

\section{Materials and methods}\label{sec:metodos}

\subsection{The three architectures}\label{sec:arq}

Table~\ref{tab:arq} lists what changes between the three architectures. Where the code of an architecture does not set a parameter, the default of the framework (Pipecat or LiveKit) or of
the service (Deepgram, OpenAI) applies. Those defaults were not changed, so each architecture
runs as its vendor ships it.

\begin{table}[htbp]
\caption{What changes between the three architectures. Everything else was the same
(Section~\ref{sec:igual})}\label{tab:arq}
\renewcommand{\arraystretch}{1.15}
\begin{tabularx}{\linewidth}{@{}P{2.3cm}YYY@{}}
\toprule
& \textbf{Pipecat} & \textbf{LiveKit Agents} & \textbf{GPT-Live}\\
\midrule
Framework & Pipecat 1.6.0 & LiveKit Agents 1.6.8 & None: an in-house bridge to the OpenAI API\\
Conversational model & \texttt{gpt-4.1-mini} & \texttt{gpt-4.1-mini} & \texttt{gpt-live-1}\\
Tool calls & \texttt{gpt-4.1-mini}, from the framework & \texttt{gpt-4.1-mini}, from the framework & GPT-Live hands each one to \texttt{gpt-4.1-mini} and can keep talking while it waits\\
Speech recognition & Deepgram \texttt{nova-3}, \texttt{multi} mode & Deepgram \texttt{nova-3}, \texttt{multi} mode & Inside the model; a transcript arrives separately\\
Speech synthesis & ElevenLabs \texttt{eleven\_flash\_v2\_5} & ElevenLabs \texttt{eleven\_flash\_v2\_5}, same voice & Inside the model, its default voice\\
Voice activity detector & Silero, 0.2~s minimum silence & Silero, 0.55~s minimum silence & Internal, not configurable\\
End-of-turn detector & Smart Turn v3, local & \texttt{TurnDetector} \texttt{v1-mini}, local & Internal, not configurable\\
Audio path in the tests & WebSocket on the same machine & Room on a LiveKit Cloud server, over the internet & Local WebSocket to the bridge; the bridge talks to the OpenAI API over the internet\\
\botrule
\end{tabularx}
\end{table}

The study compared orchestrators, not speech recognizers, so Pipecat and LiveKit used the same STT. Deepgram was chosen because it was the best integrated with both
frameworks, and because in the public comparison of real-time recognizers maintained
by the Pipecat team~\cite{sttbenchmark} its \texttt{nova-3} model was among the three
fastest and in the more accurate half of the table (consulted in September 2026). In
GPT-Live, when the model decides that a tool is needed, it opens a request that the OpenAI
service hands to \texttt{gpt-4.1-mini}; the bridge executes it on the same negotiation
engine as the other two architectures and returns the result, and the text with which
\texttt{gpt-4.1-mini} responds goes back to GPT-Live, which says it aloud. The token
counts reported by the service (median of \RCensoTokensMed{} input tokens per
request) confirmed that, as in Pipecat and LiveKit, the model that resolves the tools
receives the instructions and the conversation.

\subsection{Controlled variables}\label{sec:igual}

Everything that was not the voice technology was the same in the three architectures.
All three used a single system prompt, unchanged during the tests and containing the same trip data (Supplementary Material, Section~S1); in GPT-Live, the same text was passed as the session instructions.
All three also used the same four tools, with the same descriptions and the same negotiation engine.
The tool \texttt{proponer\_precio} (propose price) validates an amount before the agent says it and answers only ``accepted'' or ``rejected''; it rejects an amount above the maximum price (21,000 pesos, known only to this function) or one that breaks the rule of decreasing concessions.
The tool \texttt{ofrecer\_compensacion\_no\_monetaria} (offer non-monetary compensation) records a compensation from the authorized list and returns its exact wording.
The tool \texttt{registrar\_acuerdo} (register agreement) closes the call with an agreement, and \texttt{finalizar\_sin\_acuerdo} (end without agreement) closes it without one.
The measurement harness and the analysis program were also the same (Sections~\ref{sec:arnes} and~\ref{sec:estadistica}).
Finally, all runs used the same equipment: a PC with \REntSistema, an \REntCpu{} processor (\REntNucleos{} logical cores) and 16~GB of RAM, connected to the internet over Wi-Fi.

\subsection{Stimulus and scenarios}\label{sec:estimulo}

In every call, the carrier (the trucking company with which the agent negotiates) was simulated: instead of a person, the harness played a set of \RGuionLineas{} recorded lines. A person never says the
same sentence twice in the same way, and over hundreds of calls that variation is as
large as the differences to be measured. With recorded audio, each line sounds the
same in the three architectures, the instant at which the carrier's speech ends is known
rather than estimated, and the correct text for the error rate comes from the audio
that was played. One of the authors (J.~P.~M.) recorded all the lines once, with a
headset microphone. Each file went through the same steps: a gain to bring the speech
to $-18$~dBFS, conversion to 8~kHz with an anti-aliasing filter, trimming of the
silence at both ends, $\mu$-law encoding (ITU-T G.711, the telephone format) and the
end-of-speech mark, which is the zero of that line's latency. Under the speech, the harness added uniform noise at $-60$~dBFS, like the noise floor of a telephone line, and kept sending it between lines.

\begin{figure}[htbp]
\centering
\begin{fitwidth}
\begin{tikzpicture}[font=\footnotesize, node distance=0.55cm,
  caja/.style={draw, rounded corners=2pt, align=center, inner sep=4pt, text width=5.2cm},
  nota/.style={draw, gray, rounded corners=2pt, align=left, inner sep=4pt, text width=8.4cm, text=black},
  flecha/.style={-{Stealth[length=2mm]}, thick}]
\node[caja] (esc) {\textbf{Scenario} (Table~\ref{tab:escenarios})\\ a script of eight carrier lines};
\node[caja, below=of esc] (toca) {\textbf{The harness plays the line of turn $n$}\\ (a recording)};
\node[caja, below=of toca] (agente) {\textbf{The agent answers}\\ with the same instructions, trip data and tools in every scenario; it is not told which scenario it is in};
\node[caja, below=of agente] (lee) {\textbf{The harness reads what the agent said}\\ and chooses the line of turn $n+1$};
\draw[flecha] (esc) -- (toca);
\draw[flecha] (toca) -- (agente);
\draw[flecha] (agente) -- (lee);
\draw[flecha] (lee.west) -- ++(-0.45,0) |- node[pos=0.25, left, align=right, text width=1.9cm] {next turn, until turn 8 or until the agent ends the call} (toca.west);
\node[nota, right=1.1cm of lee.south east, anchor=south west] (regla) {\textbf{How the line of turn $n+1$ is chosen}\\[2pt]
Turn without branches: its only recording.\\[2pt]
Turn with branches (\RGuionTurnosConRamas{} of \RGuionBase): fixed rules, checked in order; the first that holds picks the recording. They look at what the agent said: an amount (in that response or earlier in the call), a compensation, or certain phrases.\\[6pt]
\textbf{Example: scenario 2, turn 5}\\[2pt]
If the agent has offered 18,000 or more: ``Órale, así sí me late'' (``Alright, that works for me'').\\[2pt]
Else, if it offered a compensation in that response: ``Está bien lo de la compensación, pero el número tiene que ser dieciocho mil'' (``The compensation is fine, but the number has to be eighteen thousand'').\\[2pt]
Otherwise: ``Ya casi. Nada más déjelo en 18,000 parejo'' (``Almost there. Just make it an even 18,000'').};
\draw[gray, dashed, thick] (lee.east) -- (lee.east -| regla.west);
\end{tikzpicture}
\end{fitwidth}
\caption{How a scenario is played. A scenario is the carrier's side of the call: a script of eight lines, played by the harness one per turn. The agent does not know which scenario it is in. After each response the harness reads what the agent said and, in the turns with branches, picks the next recording with fixed rules, the same for the three architectures.}\label{fig:escenario}
\end{figure}
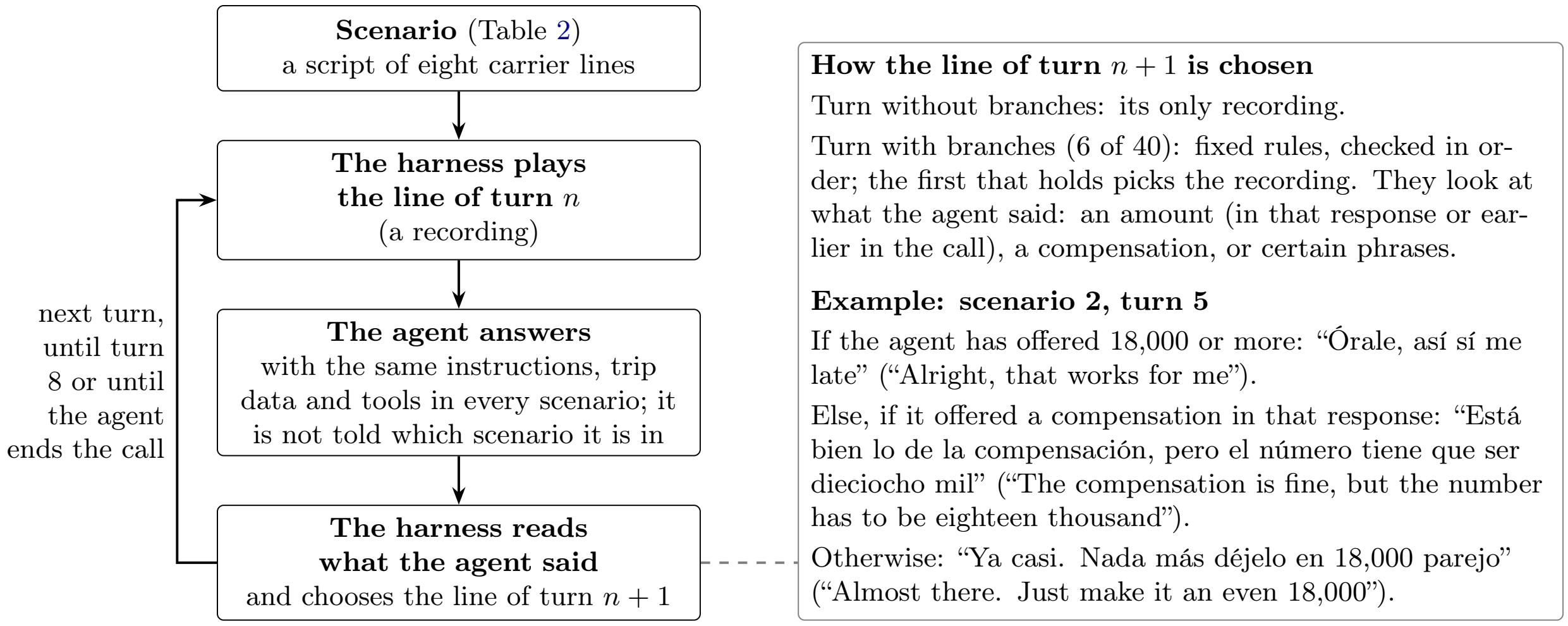

Five scenarios of eight turns each confronted the agent with a different type of carrier (Table~\ref{tab:escenarios}); all of them used the same trip data. The lines were written in the Spanish of Mexican carriers. In \RGuionTurnosConRamas{} of the
\RGuionBase{} turns, the line that played depended on what the agent said in its
previous response (for example, whether it had already offered 18,000), so that the
conversation still makes sense. The harness decided with fixed rules, the same in the
three architectures, based on what the agent said, as a carrier would. These branches added \RGuionAlt{} alternative lines to the \RGuionBase{} base lines.
A scenario is therefore only the carrier's side of the call: the agent was not told which scenario it was in, and every call started with the same instructions, trip data and tools (Section~\ref{sec:igual}).
Fig.~\ref{fig:escenario} shows how a scenario is played and how the harness chooses each line; amounts were read with the same extractor as the error on amounts (Section~\ref{sec:met-stt}), so ``dieciocho mil'' and ``18,000'' count as the same number.
All lines and rules are listed in the Supplementary Material (Section~S2).

\begin{table}[htbp]
\caption{The five scenarios, the same in the three architectures}\label{tab:escenarios}
\renewcommand{\arraystretch}{1.15}
\begin{tabularx}{\linewidth}{@{}p{0.3cm}P{3.0cm}Y@{}}
\toprule
\textbf{\#} & \textbf{Scenario} & \textbf{What it demands from the voice technology}\\
\midrule
1 & Cooperative & Baseline: a clean conversation, without pressure and without amounts; it ends in agreement.\\
2 & Moderate haggler & Several consecutive turns that call \texttt{proponer\_precio}; amounts requested again and again, and a self-correction (``Diecisiete\ldots{} no, dieciocho'', ``Seventeen\ldots{} no, eighteen'').\\
3 & Aggressive, high anchor & Amounts above the maximum (28,000, 25,000), direct pressure on the ceiling, curt tone; it ends without agreement.\\
4 & Non-monetary objection & Responses that depend on understanding the objection rather than the price: waiting time, delivery date; no amounts.\\
5 & Recognition stress test & Long amounts in words, a self-correction of an amount (18,000 to 18,500), place names and trade terms (\emph{tara}, tare; \emph{carta porte}, waybill; \emph{casetas}, toll booths).\\
\botrule
\end{tabularx}
\end{table}

\subsection{Harness and latency measurement}\label{sec:arnes}

\begin{figure}[htbp]
\centering
\begin{fitwidth}
\begin{tikzpicture}[
  font=\footnotesize,
  caja/.style={draw, rounded corners=2pt, align=center, fill=gray!8, minimum height=0.9cm, inner sep=3pt},
  poc/.style={caja, fill=blue!7, text width=3.0cm},
  datos/.style={caja, fill=green!7},
  flecha/.style={-{Stealth[length=2mm]}, thick},
  dflecha/.style={{Stealth[length=2mm]}-{Stealth[length=2mm]}, thick},
  node distance=0.45cm and 0.7cm]
\node[caja, text width=2.2cm] (banco) {\RGuionLineas{} recorded\\ carrier lines};
\node[caja, right=of banco, text width=2.7cm, fill=orange!10] (arnes) {\textbf{Harness}\\ plays the audio,\\ listens to the agent,\\ timestamps both ends};
\node[poc, right=1.0cm of arnes, yshift=1.25cm] (p1) {\textbf{Pipecat}\\ local WebSocket};
\node[poc, below=of p1] (p2) {\textbf{LiveKit Agents}\\ LiveKit Cloud server};
\node[poc, below=of p2] (p3) {\textbf{GPT-Live}\\ local bridge $\rightarrow$ API};
\node[caja, right=0.7cm of p2, text width=2.1cm] (apis) {Cloud APIs\\ Deepgram, OpenAI,\\ ElevenLabs};
\node[datos, below=1.6cm of arnes, text width=3.2cm] (reg) {Records per call:\\ marks, log and\\ two-channel audio};
\draw[flecha] (banco) -- (arnes);
\draw[dflecha] (arnes.east) -- (p1.west);
\draw[dflecha] (arnes.east) -- (p2.west);
\draw[dflecha] (arnes.east) -- (p3.west);
\draw[dflecha] (p1.east) -- (apis);
\draw[dflecha] (p2.east) -- (apis);
\draw[dflecha] (p3.east) -- (apis);
\draw[flecha] (arnes) -- (reg);
\end{tikzpicture}
\end{fitwidth}
\caption{Experimental setup. The harness replaces the carrier and is the same for the
three architectures; only the architecture it connects to changes. Double arrows are audio in both
directions.}\label{fig:montaje}
\end{figure}
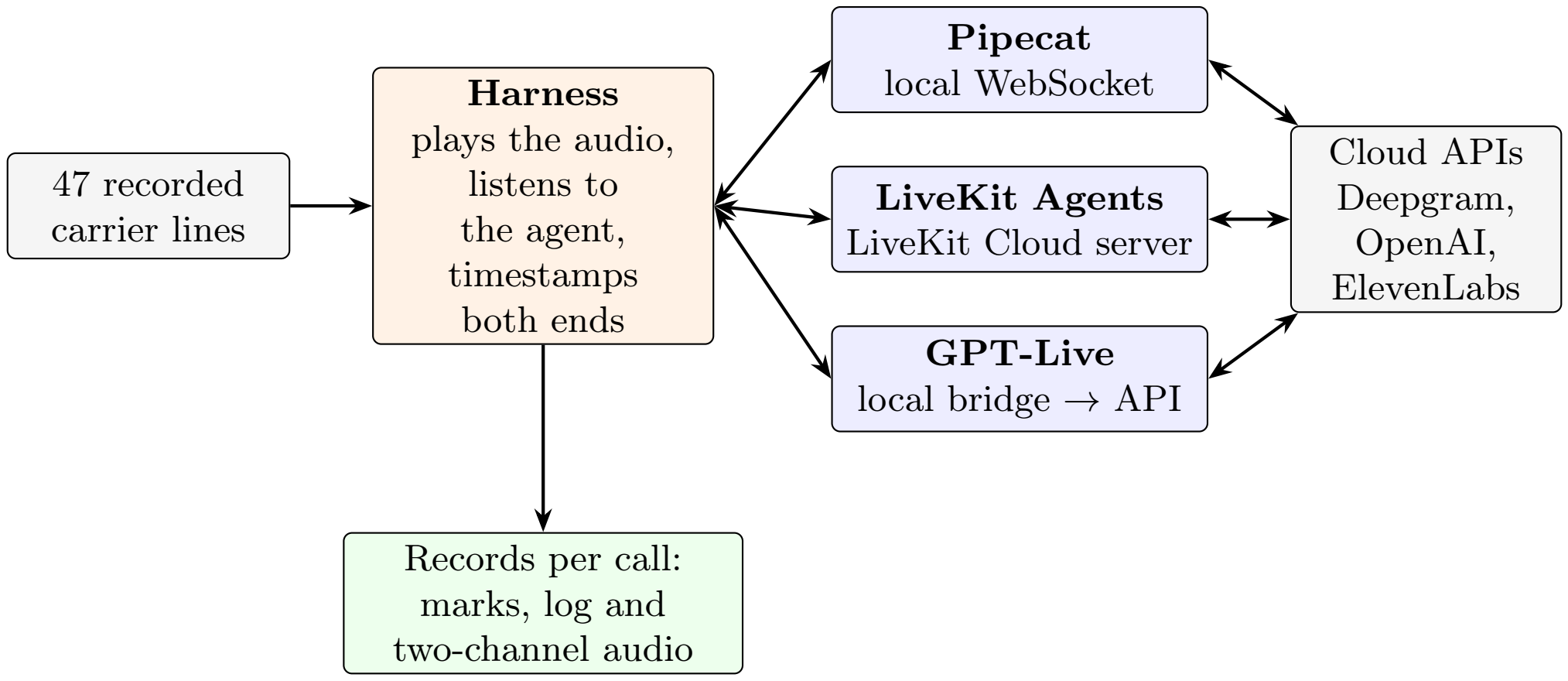

Pipecat and LiveKit compute their own latency, but that figure measures a different
interval. It starts when the framework itself detects that the carrier has stopped
speaking, which happens after the actual end of speech, and it stops when the
framework releases the first audio of its response, before that audio travels to the
carrier. The GPT-Live API does not report the latency of each response, and the
figure that OpenAI publishes for the model, a turn-taking latency of
0.8~s~\cite{gptlive}, measures the time from when the user finishes a turn until the
model starts to respond, but does not say at which point of the audio path those two
instants are taken (Section~\ref{sec:res-reloj}). A harness was therefore built: a program that played the carrier, listened to the agent and timestamped both ends on the listener's
side, with the clock of the same computer and the same rule for the three architectures
(Fig.~\ref{fig:montaje}, Table~\ref{tab:arnes}). Only the connection to each architecture changed.

\begin{table}[htbp]
\caption{Harness parameters, the same in the three architectures}\label{tab:arnes}
\renewcommand{\arraystretch}{1.15}
\begin{tabularx}{\linewidth}{@{}P{3.2cm}Y@{}}
\toprule
\textbf{Parameter} & \textbf{Value}\\
\midrule
Audio sending & 20-ms frames ($\mu$-law, 8~kHz), each sent at a fixed instant computed from the start of the call, so that delays do not accumulate.\\
End of the carrier's speech & Instant at which the first frame of the line was sent, plus the end-of-speech mark of that audio.\\
Agent speech & A 20-ms frame counts as speech if its level (RMS) reaches $-44.3$~dBFS; speech frames separated by less than 0.3~s form a run.\\
Content response & The first run that starts after the end of the carrier's audio and lasts at least 0.6~s. If it lasts less than 2.0~s, it is also discarded when its text is a backchannel (three words or fewer from a fixed list: ``ajá'', ``mm'', ``sí'', ``ok'', among others), and latency is measured to the next run that qualifies.\\
Next line & When the agent has answered and has been silent for 2.0~s; if it does not answer, after 25~s at most.\\
\botrule
\end{tabularx}
\end{table}

For each line $i$, the latency is
\begin{equation}
L_i = t^{\text{agent}}_i - t^{\text{end}}_i,
\end{equation}
where $t^{\text{end}}_i$ is the end of the carrier's speech and $t^{\text{agent}}_i$
the first audible frame of the agent's content response. The unit of analysis was the line and not the framework's turn, because each end-of-turn detector segments differently. Latency to the first sound was also reported: up to the agent's
first run of any kind, whether a filler (``un segundo\ldots'', ``one second\ldots''),
a short ``mm'' or the start of the response. The first measure tells how long the
carrier waits for content; the second, how long until the carrier knows that the agent has heard.
Both ends of the agent's response rest on the speech rule of Table~\ref{tab:arnes}.
A 20-ms frame $f$ of the agent's audio, with samples $x_{f,1},\dots,x_{f,n}$ normalized to $[-1,1]$ ($n=160$ at 8~kHz), was counted as speech when
\begin{equation}
20\log_{10}\sqrt{\frac{1}{n}\sum_{k=1}^{n}x_{f,k}^{2}}\;\geq\;-44.3~\text{dBFS},
\label{eq:frame}
\end{equation}
and consecutive speech frames separated by less than 0.3~s were joined into one run.

Observed latency includes the time the audio spends on the way, and that time is not
the same in the three architectures. It was measured in two parts: inbound transport,
from the actual end of speech to the instant the framework records as that end, and
outbound transport, from when the framework releases its audio until the
harness hears it. Both come from subtracting harness and framework marks taken with
the clock of the same computer. GPT-Live does not publish when it detects that the
carrier has finished, so in GPT-Live only outbound transport can be measured.
Writing $\tau^{\text{in}}_i$ and $\tau^{\text{out}}_i$ for the inbound and outbound transport of line $i$, and $P_i$ for the time between the two framework marks, the observed latency decomposes as
\begin{equation}
L_i=\tau^{\text{in}}_i+P_i+\tau^{\text{out}}_i ,
\label{eq:transporte}
\end{equation}
and the latency without transport is $\tilde{L}_i=L_i-\tau^{\text{in}}_i-\tau^{\text{out}}_i$, computed response by response. A line was marked with a tool if there was any tool call during the whole response,
before or after the agent started speaking. If $\mathcal{T}$ is the set of lines with a tool in a given architecture, the cost of a tool call is estimated as
\begin{equation}
\Delta_{\text{tool}}=\operatorname*{med}_{i\in\mathcal{T}}L_i-\operatorname*{med}_{i\notin\mathcal{T}}L_i .
\label{eq:herramienta}
\end{equation}
Since the LLM decides
when to use a tool, the two groups may also differ in other respects, and the cost is
approximate.

In Pipecat and LiveKit, the harness computed latency during the call. In GPT-Live it was
computed after each call, on the harness recording, with the same two ends; the turns
that the bridge log could not locate were measured directly on the recording with the
rule of Table~\ref{tab:arnes} (\RRecuperadosC{} turns, \RRecuperadosPctC\,\% of those
of GPT-Live). When measured in the same way on their recordings, Pipecat and LiveKit give practically the same
medians as when measured live (\RReconMedA{} versus \RReconObsMedA~ms, and \RReconMedB{} versus
\RReconObsMedB~ms).

For each word that the model hears or says, the GPT-Live API returns its start and
end times in milliseconds (fields \texttt{start\_ms} and \texttt{end\_ms}). We call them
GPT-Live's timestamps. They are not clock times: they indicate where in the
session audio each word lies, so they do not include the time the audio takes to
reach the model or to come back, and they advance somewhat faster than the
computer's clock (in a typical call, \RRelojDerivaMed~ms more over the
\RRelojDerivaLapsoMed~s between the first and the last line). They were therefore used only to measure intervals within a single response. With them, and with the time at
which each message from the service reaches the computer, the wait for each GPT-Live
response was split into four segments that add up to it
(Section~\ref{sec:res-reloj}). To know which word was playing when the measured speech
started, the audio received from the service was played back in sequence and in real
time from its arrival time. In addition, to find out whether the connection method
matters, separate GPT-Live sessions were opened over WebSocket, the server-side method
used in the campaign, and over WebRTC, the method that ChatGPT uses in the browser:
\RViaSesiones{} alternating sessions per method, with one-line instructions, without
tools, and the same \RViaLineas{} recorded lines in each session.

\subsection{Harness validation}\label{sec:validacion}

A measurement instrument is checked by measuring something whose value is known; because these checks establish the validity of the instrument rather than answer the study's questions, their outcomes are reported here, with the method.
The check was needed for two reasons.
First, the harness decides when the agent starts to speak from the level of its audio (Eq.~\ref{eq:frame}), so a quiet voice or a soft onset could be detected late.
Second, each architecture reaches the harness over a different audio path, so a check on one path would not hold for the others.
The known value therefore came from a simulated agent, run over the path of each architecture: a program that takes the place of the agent,
recognizes which line it is hearing and starts speaking exactly $d$~ms after that
line's end of speech. Six conditions (three delays, two speech levels and a soft
onset, in which an ``mm'' 12~dB lower precedes the sentence) were run over the five
scenarios, \RValCorridas{} calls and \RValTurnos{} turns per path, and what the
harness reported was compared with what the simulated agent did, over the actual
audio path of each architecture (Table~\ref{tab:validacion}).
For a turn $j$ in which the harness reported the latency $\hat{L}_j$, the measurement error is
\begin{equation}
e_j=\hat{L}_j-d ,
\label{eq:error}
\end{equation}
and Table~\ref{tab:validacion} gives the median of $e_j$ in each condition.

\begin{table}[htbp]
\caption{Harness validation against a simulated agent with known delay $d$. Error:
latency reported by the harness minus the actual one (ms, median of each condition).
Measured: turns with a latency, out of those run}\label{tab:validacion}
\setlength{\tabcolsep}{2.5pt}
\begin{tabular}{@{}lrrrrrr@{}}
\toprule
& \multicolumn{2}{c}{\begin{tabular}[b]{@{}c@{}}\textbf{Pipecat}\\ local WebSocket\end{tabular}} & \multicolumn{2}{c}{\begin{tabular}[b]{@{}c@{}}\textbf{LiveKit}\\ cloud server\end{tabular}} & \multicolumn{2}{c@{}}{\begin{tabular}[b]{@{}c@{}}\textbf{GPT-Live}\\ bridge\end{tabular}}\\
\cmidrule(lr){2-3}\cmidrule(lr){4-5}\cmidrule(l){6-7}
\textbf{Condition} & \textbf{Measured} & \textbf{Error} & \textbf{Measured} & \textbf{Error} & \textbf{Measured} & \textbf{Error}\\
\midrule
$d=300$~ms, $-20$~dBFS & \RValTresMedidos/\RValTresTurnos & \RValTresMediana & \RValBTresMedidos/\RValBTresTurnos & \RValBTresMediana & \RValDTresMedidos/\RValCTresTurnos & \RValDTresMediana\\
$d=1{,}000$~ms, $-20$~dBFS & \RValMilMedidos/\RValMilTurnos & \RValMilMediana & \RValBMilMedidos/\RValBMilTurnos & \RValBMilMediana & \RValDMilMedidos/\RValCMilTurnos & \RValDMilMediana\\
$d=2{,}500$~ms, $-20$~dBFS & \RValDosMilMedidos/\RValDosMilTurnos & \RValDosMilMediana & \RValBDosMilMedidos/\RValBDosMilTurnos & \RValBDosMilMediana & \RValDDosMilMedidos/\RValCDosMilTurnos & \RValDDosMilMediana\\
$d=1{,}000$~ms, $-6$~dBFS & \RValSeisMedidos/\RValSeisTurnos & \RValSeisMediana & \RValBSeisMedidos/\RValBSeisTurnos & \RValBSeisMediana & \RValDSeisMedidos/\RValCSeisTurnos & \RValDSeisMediana\\
$d=1{,}000$~ms, $-35$~dBFS & \RValTreintaMedidos/\RValTreintaTurnos & \RValTreintaMediana & \RValBTreintaMedidos/\RValBTreintaTurnos & \RValBTreintaMediana & \RValDTreintaMedidos/\RValCTreintaTurnos & \RValDTreintaMediana\\
Soft onset (``mm'') & \RValSuaveMedidos/\RValSuaveTurnos & \RValSuaveMediana & \RValBSuaveMedidos/\RValBSuaveTurnos & \RValBSuaveMediana & \RValDSuaveMedidos/\RValCSuaveTurnos & \RValDSuaveMediana\\
\botrule
\end{tabular}
\end{table}

On the Pipecat path, which does not depend on any external service, the largest error in
any turn was \RValErrMaxAbs~ms, well below the 200~ms margin used to judge differences
(Section~\ref{sec:estadistica}). The negative medians, about $-40$~ms with normal
speech, are the distance between two criteria for the start of speech and not a
measurement error. On the LiveKit path the error is the internet trip to the LiveKit
server and back (\RValBIdaNormal~ms on the way in and \RValBVueltaNormal~ms on the way
back with normal speech), which is also present in the campaign and is part of what
is called transport here. On the GPT-Live path, the direct measurement on the recording
measured every response, with the same errors as on the Pipecat path.

\FloatBarrier
\subsection{Speech recognition}\label{sec:met-stt}

The word error rate (WER) counts the substitutions $S$, deletions $D$ and insertions
$I$ needed to turn the transcript into the correct text (the reference), over the number of words in the reference $N_{\text{ref}}$. The error on amounts, $E_{\text{amt}}$, was the same count restricted to sums of money:
\begin{equation}
\mathrm{WER}=\frac{S+D+I}{N_{\text{ref}}},\qquad E_{\text{amt}}=\frac{S_{a}+D_{a}+I_{a}}{N_{a}},
\label{eq:wer}
\end{equation}
where the subscript $a$ restricts each count to the amounts and $N_a$ is the number of amounts in the reference. Amounts were singled out because in a negotiation
those are the words that cost money. WER was computed once per call: all the lines the carrier said and everything the system transcribed were joined in order, and the two texts were compared, because each system segments turns in its own way. The rate reported for each architecture adds up the errors of all its calls and divides them by the reference words of those calls. The last line was left out if the call closed less than 4~s after it finished playing, because its
transcript did not have time to arrive in full (\RWerFueraA{} calls in Pipecat,
\RWerFueraB{} in LiveKit and \RWerFueraC{} in GPT-Live). Before comparison, the texts were normalized: lower case, no accents except ``ñ'', no punctuation, and numbers written
in words or in digits treated as equal.

The error on amounts measures the transcript. To compare, with the same criterion,
what each agent understood, the self-correction in scenario~5, turn~4 was used:
\emph{``Yo andaba pensando en dieciocho mil\ldots{} no, dieciocho mil quinientos''}
(``I was thinking eighteen thousand\ldots{} no, eighteen thousand five hundred'').
J.~P.~M. read the agent's response to that line in every call, in random order and
blind to the architecture, and classified it: the agent understood 18,500 if it names or offers
that amount; it understood 18,000 if it attributes that amount to the carrier (``I
understand you are looking for 18,000''); and it cannot be told if the response only
makes a counteroffer or gives no amount.

\subsection{Interruptions and backchannels}\label{sec:met-interr}

In the main campaign, the carrier never talked over the agent. A separate scenario tested what each agent does when the carrier does. In six turns of each call, when the agent had been speaking for 1~s without a pause, the harness did one of three things: an
interruption, a line with which the carrier takes the floor (``Oiga, ¿eso
incluye las casetas?'', ``Listen, does that include the tolls?''), to which the agent
should stop and answer; a backchannel (``ajá'', ``mjm'', ``sí''), during which
it should keep talking; or a control, in which nothing plays and which gives
the rate at which the agent stops on its own at that moment. Each call had two events of each type. The agent stopped if its speech broke off with a silence of at
least 0.7~s that starts within 3~s of the event; the threshold was set before the
runs. The time to stop, the latency of the answer to an interruption, the length of
the pause and whether the agent spoke again within 8~s were also measured.

When the agent stopped at a backchannel and spoke again, what it said before and after
was taken from the recording and the resumption was classified as: continued
the sentence where it left off; repeated an offer, a clause or the start of the
sentence; said something new; took it as an answer, if it responded as if
the backchannel were the carrier's answer (for example, it took an ``ok'' as
acceptance of the price); or cannot tell. J.~P.~M. classified them, listening
to an audio clip of each resumption and reading its transcript, in random order and
without knowing the architecture; the blinding is not complete, because the GPT-Live voice is different.
In each call that the agent closed with \texttt{registrar\_acuerdo}, what the carrier
said just before was also checked: a line that accepts, a backchannel, a fragment of
a line that in full did not accept, or a line that did not yet accept. The three architectures
ran with the default interruption handling: Pipecat stops at any speech confirmed by
its detector (0.2~s), with no minimum number of words; LiveKit stops after 0.5~s of
speech, with no minimum number of words; GPT-Live has no options. Before the campaign,
a simulated agent that stops a known time after the interruption was run over the
actual path of each architecture, and the harness detected every interruption with no false
positives.

\subsection{Procedure and populations}\label{sec:poblacion}

The harness played a line, waited for the agent's response and played the next one when the agent had been silent for 2.0~s. Pipecat and LiveKit ran in one night, alternating one call of each, with the scenarios cycling from 1
to 5. Alternation spreads any change in the network or in the providers evenly across
the two, and allows each Pipecat call to be compared with its LiveKit pair, which has the
same scenario and repetition. GPT-Live ran 18 days later, with the same harness, audio
and cycle, on the public GPT-Live API with $\mu$-law at 8~kHz, the telephone format. A
first GPT-Live campaign with the alpha version of the API was replaced by this one,
because that version was withdrawn and cannot be repeated; no conclusion rests on it.

After each call, the harness checked five criteria: that every line played
completely, that the agent's recognizer produced text for every line, that there are
at least five measurable turns, that the agent answered every line and that no
response had text without a single speech frame. A call that failed any of them was excluded before its latency was examined. In addition, a combination of scenario and repetition was included only if its call passed the criteria in all three architectures, so that the three were compared on exactly the same set. A total of \RCorridasA{} calls were run
per architecture; \RRojasA{} Pipecat calls, \RRojasB{} LiveKit calls and \RRojasC{} GPT-Live calls failed
at least one criterion. Since a combination that fails in one architecture is removed from all
three, \RAnalizadasA{} calls per architecture remained, with \RNA{} measured responses in Pipecat,
\RNB{} in LiveKit and \RNC{} in GPT-Live. The interruption scenario had three repetitions of
the five scenarios per architecture, \RIntLlamadasA{} calls each, all of them included.

\subsection{Statistical analysis}\label{sec:estadistica}

Latency was summarized by its median (p50) and its 95th percentile (p95), with linear
interpolation (method~7 of Hyndman and Fan~\cite{hyndman1996}): a few very slow
responses, such as those that wait for a tool, raise the mean, while the median does
not move with them. Intervals were 95\,\% confidence intervals (95\,\% CI) obtained by
resampling (bootstrap~\cite{efron1993}) with \RReplicas{} replicates and a fixed seed,
drawing whole calls rather than individual responses, because the responses of one
call share the connection, the provider sessions and the network state. Pipecat and LiveKit
were run in pairs, so their comparison resamples pairs; GPT-Live ran on another day, so
its comparisons are unpaired.
Formally, for two architectures $A$ and $B$ with sets of calls $\mathcal{C}_A$ and $\mathcal{C}_B$, the estimated difference is
\begin{equation}
\hat{\Delta}=m(\mathcal{C}_A)-m(\mathcal{C}_B),
\label{eq:delta}
\end{equation}
where $m(\cdot)$ is the median of all the responses in a set of calls.
In replicate $b=1,\dots,n_{\text{rep}}$, with $n_{\text{rep}}=\RReplicas$, the calls of each set are drawn with replacement (pairs of calls for Pipecat versus LiveKit), giving $\hat{\Delta}^{*}_{b}$, and the 95\,\% CI is $[q_{0.025},q_{0.975}]$, the 2.5th and 97.5th percentiles of $\{\hat{\Delta}^{*}_{b}\}$.

A difference can be real and still too small to change the choice of technology.
Before any comparison was computed, a margin of 200~ms was set. It is a criterion of
this study, not a standard: no standard states from how many milliseconds a latency
difference becomes noticeable in a call. Each difference was read according to where its 95\,\% CI fell (Table~\ref{tab:veredictos}).

\begin{table}[htbp]
\caption{How the 95\,\% CI of a difference is read against the 200~ms
margin}\label{tab:veredictos}
\renewcommand{\arraystretch}{1.1}
\begin{tabularx}{\linewidth}{@{}YP{5.4cm}@{}}
\toprule
\textbf{If the 95\,\% CI of the difference\ldots} & \textbf{Verdict}\\
\midrule
includes zero and lies entirely within $\pm200$~ms & equivalent in practice\\
includes zero and extends beyond $\pm200$~ms on some side & not distinguishable from zero\\
excludes zero and lies entirely within $\pm200$~ms & different from zero, below 200~ms\\
excludes zero and lies entirely outside $\pm200$~ms & different from zero, above 200~ms\\
excludes zero and crosses $\pm200$~ms & different from zero; may or may not exceed 200~ms\\
\botrule
\end{tabularx}
\end{table}

As a second check, rank tests were applied to the median of each call ($N=\RKwN$),
because the responses of one call are not independent: Kruskal--Wallis~\cite{kruskal1952}
for the three architectures, the Wilcoxon signed-rank test~\cite{wilcoxon1945} for Pipecat versus LiveKit (paired) and the rank-sum test for GPT-Live versus each of the others (unpaired),
with Bonferroni correction~\cite{dunn1961} over the three pairs, so that a raw $p$-value is reported as $p_{\text{corr}}=\min(1,3p)$. The same tests were
applied to the WER, the character error rate and the error on amounts of each call (for amounts, only the calls whose reference contains at least one), each with its effect size: ordinal epsilon squared, $\varepsilon^2_R=H/(N-1)$, for Kruskal--Wallis, and $r=Z/\sqrt{N}$ for each pair, where $Z$ is the standardized statistic of the test and $N$ the number of calls compared (of pairs, in the paired test)~\cite{peres2026}. Because each architecture heard a somewhat different mix of
lines, owing to the branches and to the point at which each agent closes, the latency
comparisons were repeated in two ways: resampling the calls within each scenario, so that the two architectures compared have the same mix of scenarios, and giving equal weight to each of the \RLineaN{} lines heard by all three architectures. The second uses the estimator
\begin{equation}
\Delta_{\text{line}}=\frac{1}{|\mathcal{K}|}\sum_{k\in\mathcal{K}}\bigl(m_{A,k}-m_{B,k}\bigr),
\label{eq:linea}
\end{equation}
where $\mathcal{K}$ is the set of lines heard by all three architectures ($|\mathcal{K}|=\RLineaN$) and $m_{A,k}$ is the median latency of line $k$ in architecture $A$. Its interval resamples whole calls of each architecture, as in the other comparisons. In the interruption scenario,
eight hypotheses set before the campaign were tested, each with one value per call,
paired by scenario and repetition, with the signed-rank test and Bonferroni correction
(each requires $p<\alpha'=0.05/8=\RIntPruebaAlfa$). All eight hypotheses, with their variables, comparisons, $p$-values and decisions, are listed in the Supplementary Material (Section~S3). Everything else was descriptive, and counts that were not tested are flagged as such.

\subsection{Uncontrolled differences}\label{sec:asimetrias}

No comparison between three different architectures is fully symmetric, and five differences remain.
The first is transport.
Pipecat ran on the same machine as the harness, the LiveKit audio went over the internet to a LiveKit Cloud server and back, and the GPT-Live bridge talked to the OpenAI API over the internet.
This favours Pipecat, so both the observed latency and the latency without transport are reported.
The second is the silence that the voice activity detector waits for: 0.2~s in Pipecat, 0.55~s in LiveKit and unknown in GPT-Live.
It favours Pipecat in latency and LiveKit in mid-sentence cuts.
The third is the Deepgram configuration: each framework sends its own defaults (profanity filter, filler words, end-of-speech wait), and these were not equalized.
The fourth concerns tools and fillers.
GPT-Live can speak while it waits for a tool result and sometimes opens with a filler, whereas Pipecat and LiveKit usually wait in silence; latency is therefore reported with and without a tool, up to the tool result and to the first sound.
The fifth concerns days and voices: GPT-Live ran on another day, and its voice is the model's own rather than ElevenLabs'.

The code of the three architectures, of the harness, of the simulated agents and of the analysis is publicly available, together with the exported tables behind each figure (see Declarations).
Every number in Section~\ref{sec:resultados} is written into this document by the analysis program; none was typed by hand.

\section{Results}\label{sec:resultados}

\subsection{Response latency}\label{sec:latencia}

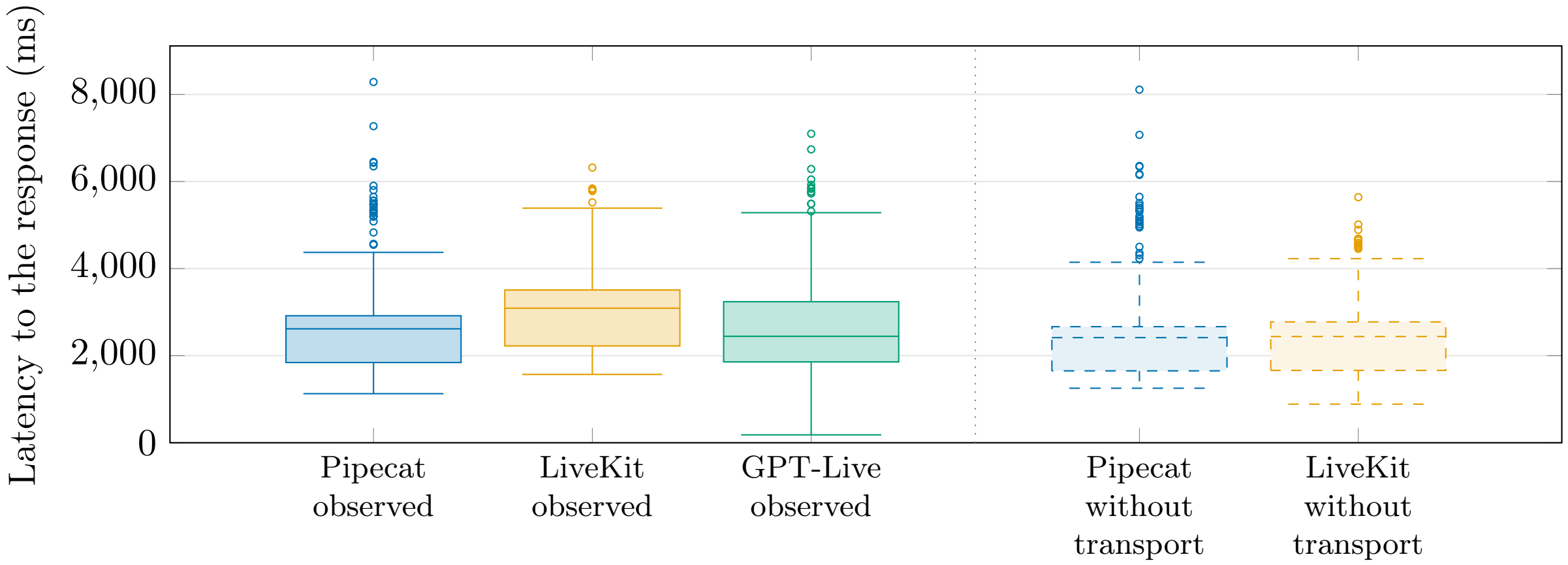
\begin{figure}[htbp]
\centering
\begin{tikzpicture}
\begin{axis}[
  width=0.95\linewidth, height=5.6cm,
  boxplot/draw direction=y, ymin=0,
  ylabel={Latency to the response (ms)},
  xtick={1,2,3,4.5,5.5},
  xticklabels={Pipecat\\observed, LiveKit\\observed, GPT-Live\\observed, Pipecat\\without\\transport, LiveKit\\without\\transport},
  xticklabel style={align=center, font=\footnotesize},
  ymajorgrids, grid style={gray!20},
  y tick label style={/pgf/number format/.cd, 1000 sep={\mathord{,}}},
]
\RCajaObsA{boxplot/draw position=1, draw=pocA, fill=pocA!25, solid, mark=o, mark size=1pt}
\RCajaObsB{boxplot/draw position=2, draw=pocB, fill=pocB!25, solid, mark=o, mark size=1pt}
\RCajaObsC{boxplot/draw position=3, draw=pocC, fill=pocC!25, solid, mark=o, mark size=1pt}
\RCajaCorrA{boxplot/draw position=4.5, draw=pocA, fill=pocA!10, dashed, mark=o, mark size=1pt, mark options={solid}}
\RCajaCorrB{boxplot/draw position=5.5, draw=pocB, fill=pocB!10, dashed, mark=o, mark size=1pt, mark options={solid}}
\draw[gray, dotted] ({axis cs:3.75,0}|-{rel axis cs:0,0}) -- ({axis cs:3.75,0}|-{rel axis cs:0,1});
\end{axis}
\end{tikzpicture}
\caption{Distribution of the latency to the content response. Boxes span the 25th to
the 75th percentile, the line is the median and the whiskers reach the last value
within 1.5 times the box height; the values beyond the whiskers are drawn as points, and no response was left out. On the right, Pipecat and LiveKit without transport; GPT-Live
has no such figure.}\label{fig:cajas}
\end{figure}

Table~\ref{tab:latencia} summarizes the latency of the three architectures and
Fig.~\ref{fig:cajas} shows its distribution. Table~\ref{tab:diferencias} gives the
differences between them, with their verdicts.
The boxes are wide for different reasons.
In Pipecat the width comes mostly from tool calls: without them, its interquartile range shrinks from \RDispIqrA{} to \RDispIqrSinHerrA~ms.
In LiveKit it comes from the end-of-turn detector, which in \RTramoFinLargoPctB\,\% of the responses judged that the carrier had not finished and waited its maximum of 2.5~s (median \RTramoFinLargoMedB~ms) instead of the usual \RTramoFinCortoMedB~ms.
In GPT-Live it comes from the time the model took to start answering according to its timestamps (segment~2 in Table~\ref{tab:reloj}), since the rest of the wait was nearly constant (Section~\ref{sec:res-reloj}).

\begin{table}[htbp]
\caption{Response latency (ms). Under each median, in brackets, its 95\,\% CI}\label{tab:latencia}
\setlength{\tabcolsep}{3pt}
\renewcommand{\arraystretch}{1.12}
\begin{tabularx}{\linewidth}{@{}Yccc@{}}
\toprule
& \textbf{Pipecat} & \textbf{LiveKit} & \textbf{GPT-Live}\\
\midrule
Calls / measured responses & \RLlamadasA{} / \RNA & \RLlamadasB{} / \RNB & \RLlamadasC{} / \RNC\\
\addlinespace
\multicolumn{4}{@{}l}{\textbf{To the content response}}\\
\quad Observed, p50 & \medci{\RObsMedA}{\RObsLoA}{\RObsHiA} & \medci{\RObsMedB}{\RObsLoB}{\RObsHiB} & \medci{\RObsMedC}{\RObsLoC}{\RObsHiC}\\
\quad Observed, p95 & \RObsNoventaycincoA & \RObsNoventaycincoB & \RObsNoventaycincoC\\
\quad Inbound transport, p50 & \RIdaA & \RIdaB & n/a\footnotemark[1]\\
\quad Outbound transport, p50 & \RVueltaA & \RVueltaB & \RVueltaC\\
\quad Without transport, p50 & \medci{\RCorrMedA}{\RCorrLoA}{\RCorrHiA} & \medci{\RCorrMedB}{\RCorrLoB}{\RCorrHiB} & n/a\footnotemark[1]\\
\quad Framework's own figure, p50 & \RPropiaA & \RPropiaB & not published\\
\addlinespace
\multicolumn{4}{@{}l}{\textbf{To the first sound}}\\
\quad Observed, p50 & \medci{\RPrimerMedA}{\RPrimerLoA}{\RPrimerHiA} & \medci{\RPrimerMedB}{\RPrimerLoB}{\RPrimerHiB} & \medci{\RPrimerMedC}{\RPrimerLoC}{\RPrimerHiC}\\
\addlinespace
\multicolumn{4}{@{}l}{\textbf{Tool calls}}\\
\quad Responses with a tool & \RRtFracA\,\% (\RRtNConA) & \RRtFracB\,\% (\RRtNConB) & \RRtFracC\,\% (\RRtNConC)\\
\quad p50 without a tool & \medci{\RRtSinA}{\RRtSinLoA}{\RRtSinHiA} & \medci{\RRtSinB}{\RRtSinLoB}{\RRtSinHiB} & \medci{\RRtSinC}{\RRtSinLoC}{\RRtSinHiC}\\
\quad p50 with a tool & \RRtConA & \RRtConB & \RRtConC\\
\quad Cost of a tool call & \medci{\RRtDeltaA}{\RRtDeltaLoA}{\RRtDeltaHiA} & \medci{\RRtDeltaB}{\RRtDeltaLoB}{\RRtDeltaHiB} & \medci{\RRtDeltaC}{\RRtDeltaLoC}{\RRtDeltaHiC}\\
\botrule
\end{tabularx}
\footnotetext[1]{GPT-Live does not publish when it detects that the carrier has
finished, so its inbound transport cannot be measured.}
\footnotetext{Latency without transport is computed response by response and its
median is then taken, so it is not the difference of the medians above. Own figure:
Pipecat's marks and LiveKit's \texttt{e2e\_latency} metric. Cost of a tool call:
median with a tool minus median without one, within the same architecture.}
\end{table}

\begin{table}[htbp]
\caption{Differences in median latency (ms), first architecture minus second. A negative value
means that the first architecture answers sooner}\label{tab:diferencias}
\setlength{\tabcolsep}{4pt}
\renewcommand{\arraystretch}{1.08}
\begin{tabularx}{\linewidth}{@{}YrcP{3.6cm}@{}}
\toprule
\textbf{Comparison} & \textbf{Difference} & \textbf{95\,\% CI} & \textbf{Verdict}\\
\midrule
\multicolumn{4}{@{}l}{\textbf{Pipecat $-$ LiveKit (paired, \RPares{} pairs)}}\\
\quad All responses & \RDifUnoDos & [\RDifLoUnoDos, \RDifHiUnoDos] & \verdict{\RDifVeredictoUnoDos}\\
\quad Without transport & \RDifCorrUnoDos & [\RDifCorrLoUnoDos, \RDifCorrHiUnoDos] & \verdict{\RDifCorrVeredictoUnoDos}\\
\addlinespace
\multicolumn{4}{@{}l}{\textbf{Pipecat $-$ GPT-Live (unpaired)}}\\
\quad All responses & \RDifUnoTres & [\RDifLoUnoTres, \RDifHiUnoTres] & \verdict{\RDifVeredictoUnoTres}\\
\quad Without outbound transport\footnotemark[1] & \RDifSinSalidaUnoTres & [\RDifSinSalidaLoUnoTres, \RDifSinSalidaHiUnoTres] & \verdict{\RDifSinSalidaVeredictoUnoTres}\\
\quad Resampling within each scenario & \RMezclaDifUnoTres & [\RMezclaDifLoUnoTres, \RMezclaDifHiUnoTres] & \verdict{\RMezclaDifVeredictoUnoTres}\\
\quad Equal weight per line & \RLineaDifUnoTres & [\RLineaDifLoUnoTres, \RLineaDifHiUnoTres] & \verdict{\RLineaDifVeredictoUnoTres}\\
\quad Up to the tool result & \RDifHastaResUnoTres & [\RDifHastaResLoUnoTres, \RDifHastaResHiUnoTres] & \verdict{\RDifHastaResVeredictoUnoTres}\\
\quad To the first sound & \RPrimerDifUnoTresIc & [\RPrimerDifLoUnoTres, \RPrimerDifHiUnoTres] & \verdict{\RPrimerDifVeredictoUnoTres}\\
\addlinespace
\multicolumn{4}{@{}l}{\textbf{LiveKit $-$ GPT-Live (unpaired)}}\\
\quad All responses & \RDifDosTres & [\RDifLoDosTres, \RDifHiDosTres] & \verdict{\RDifVeredictoDosTres}\\
\quad Equal weight per line & \RLineaDifDosTres & [\RLineaDifLoDosTres, \RLineaDifHiDosTres] & \verdict{\RLineaDifVeredictoDosTres}\\
\botrule
\end{tabularx}
\footnotetext[1]{The median of its outbound transport is subtracted from each architecture; the
inbound transport of GPT-Live cannot be measured.}
\footnotetext{Up to the tool result: in the three architectures, responses that started before
the result of a price or compensation tool arrived are measured up to that result.
Equal weight per line: the median of each line in each architecture, averaged over the
\RLineaN{} lines heard by all three.}
\end{table}

\begin{figure}[htbp]
\centering
\begin{tikzpicture}
\begin{axis}[
  width=0.74\linewidth, height=7.8cm,
  xbar, bar width=4.5pt, xmin=0,
  xlabel={Median (ms)},
  ytick={-1,...,7},
  yticklabels={GPT-Live: outside its timestamps, GPT-Live: per its timestamps, Outbound transport, Synthesis (TTS), Tool (cost), Language model, Recognition (STT), End of turn, Inbound transport},
  yticklabel style={font=\footnotesize},
  xmajorgrids, grid style={gray!20},
  x tick label style={/pgf/number format/.cd, 1000 sep={\mathord{,}}},
  legend style={at={(0.98,0.98)}, anchor=north east, font=\footnotesize},
  reverse legend,
  enlarge y limits=0.06,
]
\addplot[draw=pocC, fill=pocC!60, bar shift=-5pt] coordinates {\RCoordCapasC};
\addplot[draw=pocB, fill=pocB!60, bar shift=0pt] coordinates {\RCoordCapasB};
\addplot[draw=pocA, fill=pocA!60, bar shift=5pt] coordinates {\RCoordCapasA};
\addplot[draw=pocC, fill=pocC!60, bar shift=0pt, forget plot] coordinates {(\RRelojModeloValor,0) (\RRelojRestoValor,-1)};
\legend{GPT-Live, LiveKit, Pipecat}
\draw[gray!60, dashed] ({rel axis cs:0,0}|-{axis cs:0,0.5}) -- ({rel axis cs:1,0}|-{axis cs:0,0.5});
\end{axis}
\end{tikzpicture}
\caption{Where the time of a response goes: median of each stage, in ms. In the
cascades, the stages come from the internal marks of Pipecat and LiveKit (end of turn:
from the end of speech to the decision that the turn has ended; recognition: to the
final transcript; language model: to the first chunk of the response, without a tool;
synthesis: to the first chunk of audio), and transport and the cost of a tool call come
from the harness. GPT-Live does not separate those stages: its two rows are what it
took according to its own timestamps and what those timestamps do not include
(Section~\ref{sec:res-reloj}). The bars cannot be added up, because some stages run at
the same time (the recognizer transcribes while the detector decides whether the
carrier has finished) and each was measured separately.}\label{fig:capas}
\end{figure}
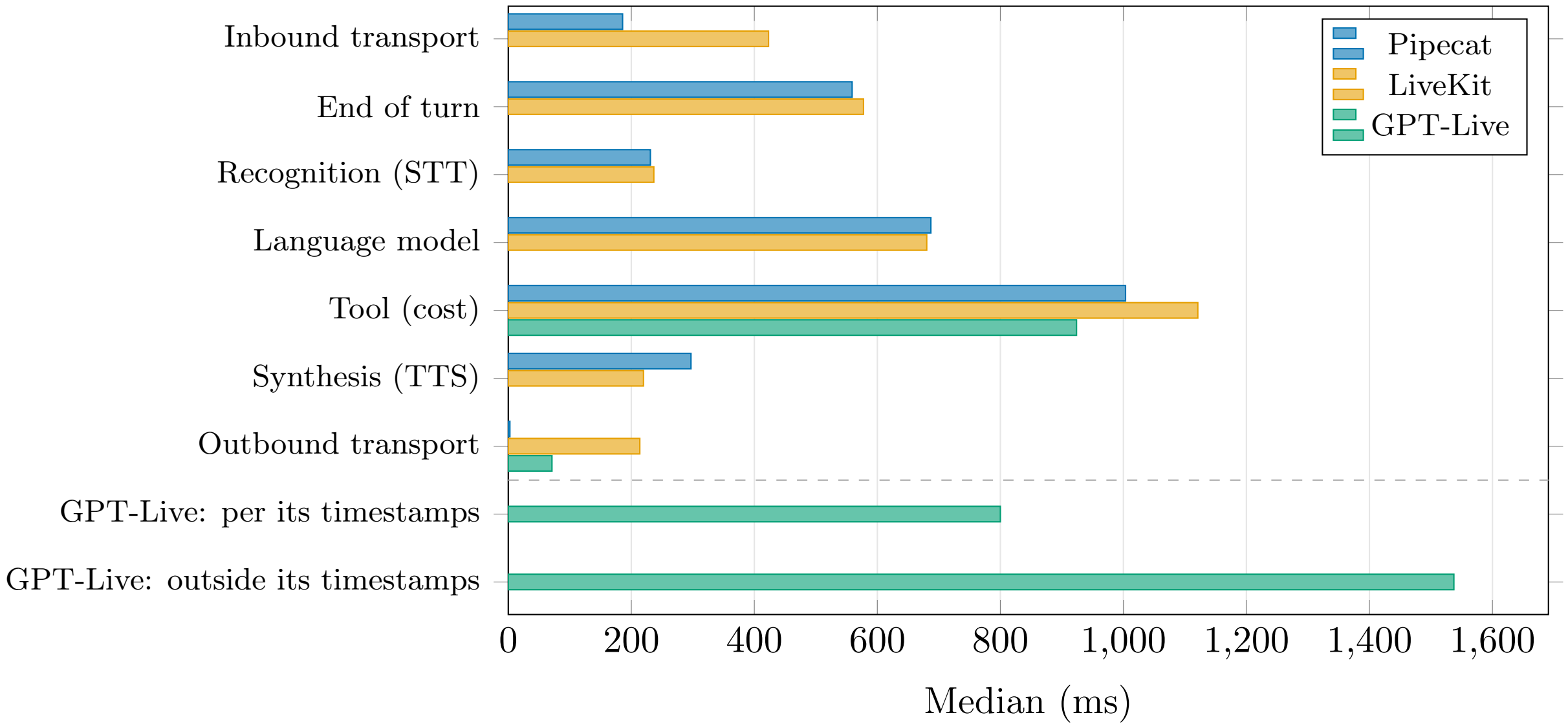

\paragraph{LiveKit was the slowest, because of its transport.} LiveKit was
\RDifAbsUnoDos~ms behind Pipecat and \RDifAbsDosTres~ms behind GPT-Live, with intervals
entirely outside the 200~ms margin, and it remains the slowest under every way of
counting. Its two transports add up to about \RTransporteB~ms (\RIdaB{} inbound and
\RVueltaB{} outbound), against \RTransporteA~ms in Pipecat (Fig.~\ref{fig:capas}).
Without transport, Pipecat and LiveKit are \verdict{\RDifCorrVeredictoUnoDos}
(\RDifCorrUnoDos~ms). The language model and the recognizer took almost the same time
in both cascades (\REtapaLlmA{} versus \REtapaLlmB~ms to the first LLM chunk, and
\REtapaSttA{} versus \REtapaSttB~ms to the final transcript): neither framework sped
up or slowed down its providers.
The recognizer's time agrees with the public benchmark maintained by the Pipecat team, which measures it in the same way, from the end of speech to the final transcript, although on a different set of recordings, and reports a median of 247~ms for \texttt{nova-3}~\cite{sttbenchmark}.
Recognition does not lengthen the wait, because it runs while the end-of-turn detector decides, and it finished before the detector in \RTramoSttAntesPctA\,\% of the responses in Pipecat and \RTramoSttAntesPctB\,\% in LiveKit (Fig.~\ref{fig:reloj}).

\paragraph{There is no clear winner between Pipecat and GPT-Live.} Taking the median
of all responses, GPT-Live is \RDifAbsUnoTres~ms ahead of Pipecat, but the interval includes
zero (\verdict{\RDifVeredictoUnoTres}). Two ways of counting do exclude zero: removing
outbound transport, which favours Pipecat (\RVueltaA{} versus \RVueltaC~ms), and
resampling within each scenario. The others give no winner: with equal weight for each
line of the script the two are \verdict{\RLineaDifVeredictoUnoTres}, and Pipecat was
faster on \RLineaMasRapidaUnoTres{} of the \RLineaN{} lines; up to the tool result the
difference is \verdict{\RDifHastaResVeredictoUnoTres}; and the rank test on the median
of each call does not separate them either ($p=\RWPBonfUnoTres$, corrected). GPT-Live's
advantage in the overall median comes from the lines that were heard most often, not
from all of them. By scenario, GPT-Live had the lowest median in scenarios~1 and~4, the
two in which the carrier says no amounts, and Pipecat in scenarios~2, 3 and~5.

\subsection{Latency reported by the frameworks}\label{sec:res-propia}

With their own metrics, Pipecat reported a median of \RPropiaA~ms when the listener
waited \RObsMedA~ms, \RPropiaFaltaA~ms (\RPropiaFaltaPctA\,\%) less, and LiveKit
\RPropiaB~ms against \RObsMedB~ms, \RPropiaFaltaB~ms (\RPropiaFaltaPctB\,\%) less.
What is missing is mostly the two transports: the heard latency minus them is
\RCorrMedA~ms in Pipecat and \RCorrMedB~ms in LiveKit, close to what each framework
reports. With the frameworks' own figures the two would have looked almost the same
(\RPropiaA{} versus \RPropiaB~ms), and the \RDifAbsUnoDos~ms that, as deployed,
LiveKit made the carrier wait beyond Pipecat would have remained hidden.
Fig.~\ref{fig:reloj} shows where the difference lies in a typical response without a tool: each framework counts the end of turn, the language model and synthesis, and leaves out the two transports.

\subsection{The published GPT-Live figure and the measured wait}\label{sec:res-reloj}

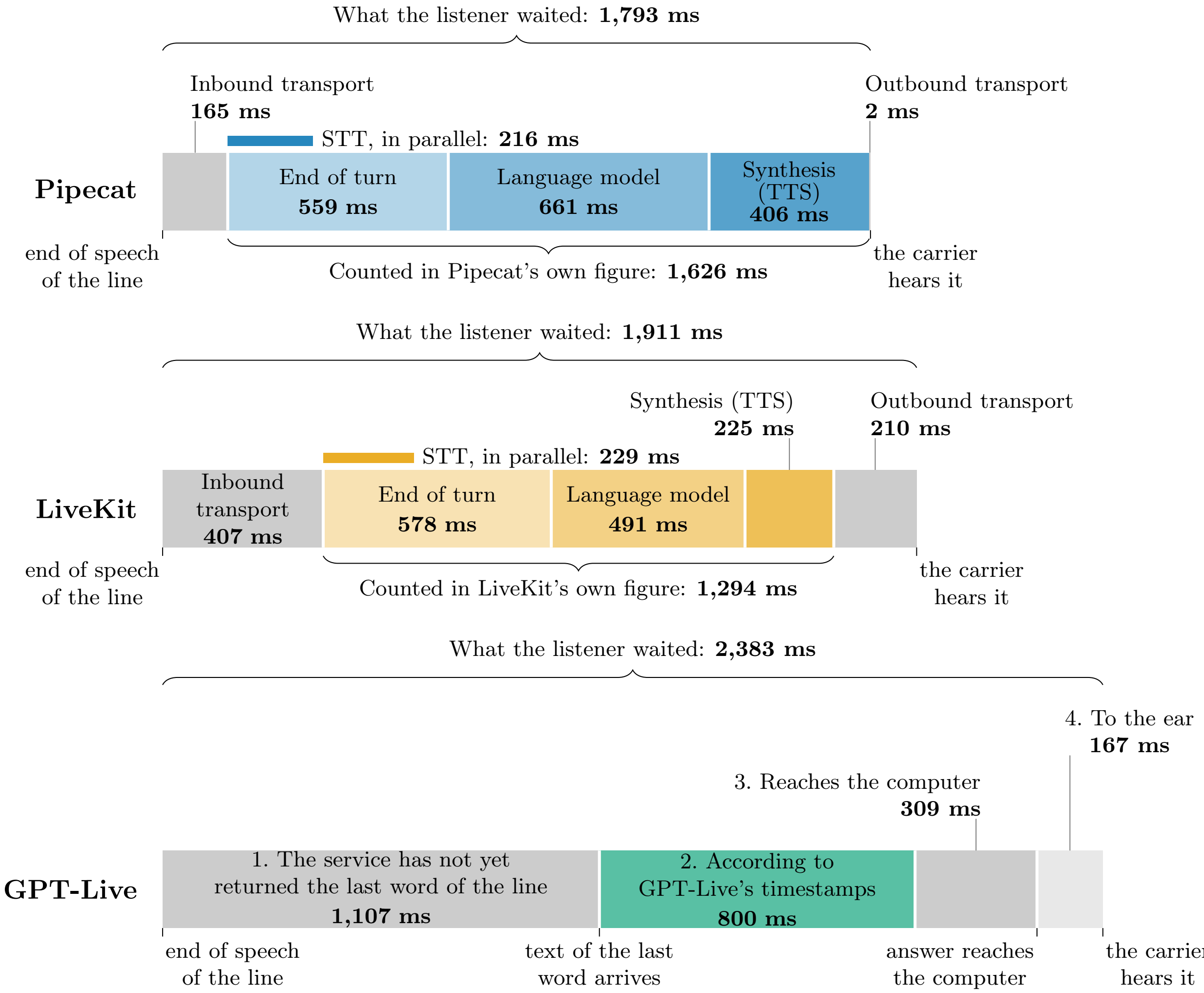
\begin{figure}[htbp]
\centering
\begin{fitwidth}
\begin{tikzpicture}[x=5.6cm, y=1cm, font=\footnotesize,
  rotulo/.style={align=center, inner sep=1pt, execute at begin node={\hyphenpenalty=10000\exhyphenpenalty=10000}},
  nombre/.style={anchor=east, font=\normalsize\bfseries}]
\begin{scope}[yshift=9.9cm]
\pgfmathsetmacro{\pa}{\RTramoEjIdaA}
\pgfmathsetmacro{\pb}{\pa+\RTramoEjFinA}
\pgfmathsetmacro{\pc}{\pb+\RTramoEjLlmA}
\pgfmathsetmacro{\pd}{\pc+\RTramoEjTtsA}
\pgfmathsetmacro{\pe}{\pd+\RTramoEjVueltaA}
\node[nombre] at (-0.04,0.55) {Pipecat};
\fill[gray!40] (0,0) rectangle (\pa,1.1);
\fill[pocA!30] (\pa,0) rectangle (\pb,1.1);
\fill[pocA!48] (\pb,0) rectangle (\pc,1.1);
\fill[pocA!66] (\pc,0) rectangle (\pd,1.1);
\fill[gray!40] (\pd,0) rectangle (\pe,1.1);
\draw[white, line width=1.2pt] (\pa,0) -- (\pa,1.1) (\pb,0) -- (\pb,1.1) (\pc,0) -- (\pc,1.1);
\node[rotulo, text width=3.0cm] at ({(\pa+\pb)/2},0.55) {End of turn\\[1pt] \textbf{\RTramoEjFinMsA~ms}};
\node[rotulo, text width=3.4cm] at ({(\pb+\pc)/2},0.55) {Language model\\[1pt] \textbf{\RTramoEjLlmMsA~ms}};
\node[rotulo, text width=2.1cm] at ({(\pc+\pd)/2},0.55) {Synthesis\\[-2pt] (TTS)\\[-2pt] \textbf{\RTramoEjTtsMsA~ms}};
\draw[gray] ({\pa/2},1.1) -- ({\pa/2},1.55);
\node[rotulo, anchor=south west, align=left] at ({\pa/2-0.02},1.55) {Inbound transport\\ \textbf{\RTramoEjIdaMsA~ms}};
\draw[gray] ({(\pd+\pe)/2},1.1) -- ({(\pd+\pe)/2},1.55);
\node[rotulo, anchor=south west, align=left] at ({(\pd+\pe)/2-0.02},1.55) {Outbound transport\\ \textbf{\RTramoEjVueltaMsA~ms}};
\draw (0,0) -- (0,-0.12) (\pe,0) -- (\pe,-0.12);
\node[rotulo, anchor=north east] at (0,-0.16) {end of speech\\ of the line};
\node[rotulo, anchor=north west] at (\pe,-0.16) {the carrier\\ hears it};
\draw[decorate, decoration={brace, mirror, amplitude=6pt}] (\pa,-0.12) -- node[below=6pt] {Counted in Pipecat's own figure: \textbf{\RTramoEjPropiaMsA~ms}} (\pd,-0.12);
\draw[decorate, decoration={brace, amplitude=6pt}] (0,2.55) -- node[above=6pt] {What the listener waited: \textbf{\RTramoEjTotalMsA~ms}} (\pe,2.55);
\fill[pocA!85] (\pa,1.2) rectangle ({\pa+\RTramoEjSttA},1.34);
\node[rotulo, anchor=west] at ({\pa+\RTramoEjSttA+0.015},1.27) {STT, in parallel: \textbf{\RTramoEjSttMsA~ms}};
\end{scope}
\begin{scope}[yshift=5.4cm]
\pgfmathsetmacro{\la}{\RTramoEjIdaB}
\pgfmathsetmacro{\lb}{\la+\RTramoEjFinB}
\pgfmathsetmacro{\lc}{\lb+\RTramoEjLlmB}
\pgfmathsetmacro{\ld}{\lc+\RTramoEjTtsB}
\pgfmathsetmacro{\le}{\ld+\RTramoEjVueltaB}
\node[nombre] at (-0.04,0.55) {LiveKit};
\fill[gray!40] (0,0) rectangle (\la,1.1);
\fill[pocB!30] (\la,0) rectangle (\lb,1.1);
\fill[pocB!48] (\lb,0) rectangle (\lc,1.1);
\fill[pocB!66] (\lc,0) rectangle (\ld,1.1);
\fill[gray!40] (\ld,0) rectangle (\le,1.1);
\draw[white, line width=1.2pt] (\la,0) -- (\la,1.1) (\lb,0) -- (\lb,1.1) (\lc,0) -- (\lc,1.1) (\ld,0) -- (\ld,1.1);
\node[rotulo, text width=2.2cm] at ({\la/2},0.55) {Inbound\\ transport\\ \textbf{\RTramoEjIdaMsB~ms}};
\node[rotulo, text width=3.0cm] at ({(\la+\lb)/2},0.55) {End of turn\\[1pt] \textbf{\RTramoEjFinMsB~ms}};
\node[rotulo, text width=2.6cm] at ({(\lb+\lc)/2},0.55) {Language model\\[1pt] \textbf{\RTramoEjLlmMsB~ms}};
\draw[gray] ({(\lc+\ld)/2},1.1) -- ({(\lc+\ld)/2},1.55);
\node[rotulo, anchor=south east, align=right] at ({(\lc+\ld)/2+0.02},1.55) {Synthesis (TTS)\\ \textbf{\RTramoEjTtsMsB~ms}};
\draw[gray] ({(\ld+\le)/2},1.1) -- ({(\ld+\le)/2},1.55);
\node[rotulo, anchor=south west, align=left] at ({(\ld+\le)/2-0.02},1.55) {Outbound transport\\ \textbf{\RTramoEjVueltaMsB~ms}};
\draw (0,0) -- (0,-0.12) (\le,0) -- (\le,-0.12);
\node[rotulo, anchor=north east] at (0,-0.16) {end of speech\\ of the line};
\node[rotulo, anchor=north west] at (\le,-0.16) {the carrier\\ hears it};
\draw[decorate, decoration={brace, mirror, amplitude=6pt}] (\la,-0.12) -- node[below=6pt] {Counted in LiveKit's own figure: \textbf{\RTramoEjPropiaMsB~ms}} (\ld,-0.12);
\draw[decorate, decoration={brace, amplitude=6pt}] (0,2.55) -- node[above=6pt] {What the listener waited: \textbf{\RTramoEjTotalMsB~ms}} (\le,2.55);
\fill[pocB!85] (\la,1.2) rectangle ({\la+\RTramoEjSttB},1.34);
\node[rotulo, anchor=west] at ({\la+\RTramoEjSttB+0.015},1.27) {STT, in parallel: \textbf{\RTramoEjSttMsB~ms}};
\end{scope}
\pgfmathsetmacro{\fa}{\RRelojEjUltima}
\pgfmathsetmacro{\fb}{\fa+\RRelojEjModelo}
\pgfmathsetmacro{\fc}{\fb+\RRelojEjSalida}
\pgfmathsetmacro{\fd}{\fc+\RRelojEjEntrega}
\node[nombre] at (-0.04,0.55) {GPT-Live};
\fill[gray!40] (0,0) rectangle (\fa,1.1);
\fill[pocC!65] (\fa,0) rectangle (\fb,1.1);
\fill[gray!40] (\fb,0) rectangle (\fc,1.1);
\fill[gray!18] (\fc,0) rectangle (\fd,1.1);
\draw[white, line width=1.2pt] (\fa,0) -- (\fa,1.1) (\fb,0) -- (\fb,1.1) (\fc,0) -- (\fc,1.1);
\node[rotulo, text width=5.6cm] at ({\fa/2},0.55) {1.~The service has not yet returned the last word of the line\\[1pt] \textbf{\RRelojEjUltimaMs~ms}};
\node[rotulo, text width=3.9cm] at ({(\fa+\fb)/2},0.55) {2.~According to GPT-Live's timestamps\\[1pt] \textbf{\RRelojEjModeloMs~ms}};
\draw[gray] ({(\fb+\fc)/2},1.1) -- ({(\fb+\fc)/2},1.55);
\node[rotulo, anchor=south east, align=right] at ({(\fb+\fc)/2+0.02},1.55) {3.~Reaches the computer\\ \textbf{\RRelojEjSalidaMs~ms}};
\draw[gray] ({(\fc+\fd)/2},1.1) -- ({(\fc+\fd)/2},2.45);
\node[rotulo, anchor=south west] at ({(\fc+\fd)/2-0.02},2.45) {4.~To the ear\\ \textbf{\RRelojEjEntregaMs~ms}};
\foreach \x in {0,\fa,\fc,\fd} {\draw (\x,0) -- (\x,-0.12);}
\node[rotulo, anchor=north west] at (0,-0.16) {end of speech\\ of the line};
\node[rotulo, anchor=north] at (\fa,-0.16) {text of the last\\ word arrives};
\node[rotulo, anchor=north east] at (\fc,-0.16) {answer reaches\\ the computer};
\node[rotulo, anchor=north west] at (\fd,-0.16) {the carrier\\ hears it};
\draw[decorate, decoration={brace, amplitude=6pt}] (0,3.45) -- node[above=6pt] {What the listener waited: \textbf{\RRelojEjTotalMs~ms}} (\fd,3.45);
\end{tikzpicture}
\end{fitwidth}
\caption{The segments of a typical response of each architecture, without a tool and
with each segment close to its median, on the same time scale. In colour, what each
architecture's own figure covers: in Pipecat and LiveKit, the latency that the framework
reports (end of turn, language model and synthesis); in GPT-Live, in green, the segment
measured by its timestamps, which resembles the figure published by OpenAI. In grey,
what those figures leave out. In the cascades, the thin bar above the end of turn is
speech recognition (STT): it starts at the same end of speech and runs while the
detector decides whether the turn has ended, so it does not lengthen the response.
In LiveKit the language model starts before the end of
the turn is confirmed (preemptive generation, on by default) and synthesis waits for
the confirmation, so its bar shows only the part of the language model that comes
after it; in the median, \RTramoTraslapeMedB~ms of the language model ran during the
end-of-turn wait.}\label{fig:reloj}
\end{figure}

OpenAI announces a turn-taking latency of 0.798~s for GPT-Live, counted from when the
user finishes a turn until the model starts to respond~\cite{gptlive}; the announcement
does not say at which point of the audio path those two instants are taken. In the
campaign, the listener waited a median of \RObsMedC~ms. The two figures do not contradict each other: they measure
different segments of the same path, which runs from the end of the carrier's speech
to the moment the carrier hears the answer. Table~\ref{tab:reloj} gives the median of
each segment for the \RRelojN{} GPT-Live responses in which the word that started the
speech could be located (out of \RNC), and Fig.~\ref{fig:reloj} draws a typical
response next to a typical response of each cascade.
The wait divides into four consecutive segments, numbered as in the table.
The first, and the longest, runs from the end of speech until the OpenAI service returns the text of the last word of the line (\RRelojUltimaMed~ms).
The second, measured with GPT-Live's timestamps, runs from the end of that last word to the start of the answer (\RRelojModeloMed~ms; \RRelojModeloSinHerrMed~ms without a tool and \RRelojModeloConHerrMed~ms with one); it is the time the model took to answer, and the segment that resembles the published figure.
The third is the time the answer took to reach the study computer beyond the two previous segments (\RRelojSalidaMed~ms).
The fourth runs from the computer to the carrier's ear, including the playback queue (\RRelojEntregaMed~ms).

Segments 1 and 3 take place in the network and in the OpenAI service and add up to
\RRelojServicioMed~ms, almost the same in every response (interquartile range from
\RRelojServicioQa{} to \RRelojServicioQb~ms) and with or without a tool. The network
is a small part of this: opening a connection to the entry server of the API takes
about \RRedMed~ms round trip. GPT-Live's timestamps do not
include these two segments, because they indicate where each word lies in the audio,
not when the service processed it (Section~\ref{sec:arnes}). This is why the published
figure resembles segment 2 and not the whole wait.

\begin{table}[htbp]
\caption{The four segments of the wait for a GPT-Live response. Median [25th percentile,
75th percentile], in ms}\label{tab:reloj}
\renewcommand{\arraystretch}{1.12}
\begin{tabularx}{\linewidth}{@{}Yccc@{}}
\toprule
& \textbf{All} & \textbf{Without a tool} & \textbf{With a tool}\\
\midrule
Responses & \RRelojN & \RRelojNSinHerr & \RRelojNConHerr\\
\addlinespace
1. Until the service returns the text of the last word & \RRelojUltimaMed{} [\RRelojUltimaQa, \RRelojUltimaQb] & \RRelojUltimaSinHerrMed & \RRelojUltimaConHerrMed\\
2. According to GPT-Live's timestamps & \RRelojModeloMed{} [\RRelojModeloQa, \RRelojModeloQb] & \RRelojModeloSinHerrMed & \RRelojModeloConHerrMed\\
3. Until the answer reaches the computer, beyond 1 and 2 & \RRelojSalidaMed{} [\RRelojSalidaQa, \RRelojSalidaQb] & \RRelojSalidaSinHerrMed & \RRelojSalidaConHerrMed\\
4. From the computer to the ear & \RRelojEntregaMed{} [\RRelojEntregaQa, \RRelojEntregaQb] & \RRelojEntregaSinHerrMed & \RRelojEntregaConHerrMed\\
\addlinespace
Measured wait & \RRelojLatMed{} [\RRelojLatQa, \RRelojLatQb] & \RRelojLatSinHerrMed & \RRelojLatConHerrMed\\
\quad in the network and the OpenAI service (1 and 3) & \RRelojServicioMed{} [\RRelojServicioQa, \RRelojServicioQb] & \RRelojServicioSinHerrMed & \RRelojServicioConHerrMed\\
\quad outside GPT-Live's timestamps (1, 3 and 4) & \RRelojRestoMed{} [\RRelojRestoQa, \RRelojRestoQb] & \RRelojRestoSinHerrMed & \RRelojRestoConHerrMed\\
\botrule
\end{tabularx}
\footnotetext{In each response, the four segments add up to the measured wait. Their
medians, however, do not add up to the median of the wait (\RRelojSumaMedMs{} versus
\RRelojLatMed~ms), because each median comes from a different response. Segment 3:
from the arrival of the text of the last word to the arrival of the word that starts
the answer, minus segment~2. Segment 4: includes the playback queue.}
\end{table}

The separate sessions confirm that segments 1 and 3 do not depend on what the agent
says. With one-line instructions and no tools, the model began to answer almost as the
line ended, according to its timestamps (median of \RViaWsModeloMed~ms over WebSocket
and \RViaRtcModeloMed~ms over WebRTC). Even so, its first audio arrived \RViaWsMed{}
and \RViaRtcMed~ms after the end of speech: the part not covered by its timestamps was
\RViaWsFueraMed{} and \RViaRtcFueraMed~ms (\RViaWsN{} lines per method), almost the
same as segments 1 and 3 in the campaign. Those sessions seem faster only because the
model hardly took any time to decide what to say. Connecting over WebRTC did not
shorten the path.

From the outside it cannot be determined in which part of the service those
\RRelojServicioMed~ms are spent: receiving the audio, transcription, waiting for
computation or encoding the voice. The split between segments 1 and 3 also assumes
that the service returns the text of a word as soon as the model hears it; if it
returns it later, part of segment 1 would belong to segment 3, and the sum of the two
would not change.

\subsection{Tool calls}\label{sec:res-herr}

In Pipecat and LiveKit, about two out of three responses included a tool call
(\RRtFracA\,\% in Pipecat and \RRtFracB\,\% in LiveKit), and each call added about one second
(Table~\ref{tab:latencia}): it is the most expensive layer of the wait in both
cascades (Fig.~\ref{fig:capas}). Without a tool, Pipecat was the fastest (\RRtSinA~ms),
and GPT-Live (\RRtSinC~ms) was close to LiveKit (\RRtSinB~ms). The overall median of GPT-Live is
low mainly because it uses tools in fewer responses (\RRtFracC\,\%). It is not that it
uses them later: counting only the calls made before the agent starts to speak lowers
the proportion to \RRtFracPrevA\,\% in Pipecat and \RRtFracPrevB\,\% in LiveKit, but leaves
GPT-Live at \RRtFracPrevC\,\%. In \RHerrHablo{} of the \RHerrN{} GPT-Live responses that waited
for a price or compensation result (\RHerrFrac\,\%), the agent had already started to
speak before it had the result, a median of \RHerrMs~ms earlier. Measured up to the
result in all three, the medians are \RHastaResMedA~ms in Pipecat, \RHastaResMedB~ms in
LiveKit and \RHastaResMedC~ms in GPT-Live.

\subsection{Latency to the first sound}\label{sec:res-primer}

In most responses the first sound is the response. A short preceding sound
appeared in \RPrimerFracC\,\% of the GPT-Live responses, \RPrimerFracA\,\% of the Pipecat
responses and \RPrimerFracB\,\% of the LiveKit responses, but it is not the same thing in
all of them. In Pipecat it is not a filler: it is the start of the response itself
(``Perfecto, Carlos.'', ``Perfect, Carlos.''), followed by a short pause and the rest;
since it lasts less than 0.6~s, the harness measures to what follows, which counts
against Pipecat. In GPT-Live it is a filler (``mm'', ``déjame revisar'', ``let me check''),
sometimes while it waits for a tool. Measured to the first sound, GPT-Live is
\RPrimerDifUnoTresIc~ms ahead of Pipecat (\verdict{\RPrimerDifVeredictoUnoTres}): the
carrier hears the agent sooner, even though the content does not arrive sooner.

\subsection{Speech recognition}\label{sec:res-stt}

\begin{table}[htbp]
\caption{Recognition errors by architecture (\%), computed per whole call, and the tests
between architectures. Under each error, in brackets, its 95\,\% CI}\label{tab:stt}
\setlength{\tabcolsep}{5pt}
\begin{tabular}{@{}lccc@{}}
\toprule
& \textbf{WER} & \textbf{Character error rate} & \textbf{Error on amounts}\\
\midrule
Pipecat & \medci{\RWerA}{\RWerLoA}{\RWerHiA} & \medci{\RCerA}{\RCerLoA}{\RCerHiA} & \medci{\RMontosA}{\RMontosLoA}{\RMontosHiA}\\
LiveKit & \medci{\RWerB}{\RWerLoB}{\RWerHiB} & \medci{\RCerB}{\RCerLoB}{\RCerHiB} & \medci{\RMontosB}{\RMontosLoB}{\RMontosHiB}\\
GPT-Live & \medci{\RWerC}{\RWerLoC}{\RWerHiC} & \medci{\RCerC}{\RCerLoC}{\RCerHiC} & \medci{\RMontosC}{\RMontosLoC}{\RMontosHiC}\\
\addlinespace
\multicolumn{4}{@{}l}{\textbf{Tests on the error of each call}}\\
\quad Calls per architecture & \RWerPruebasA & \RCerPruebasA & \RMontosPruebasA\footnotemark[1]\\
\quad Kruskal--Wallis, $p$ ($\varepsilon^2_R$) & \RWerKwP{} (\RWerKwEpsilon) & \RCerKwP{} (\RCerKwEpsilon) & \RMontosKwP{} (\RMontosKwEpsilon)\\
\quad Pipecat vs LiveKit, $p$ ($r$) & \RWerWPBonfUnoDos{} (\RWerEfectoUnoDos) & \RCerWPBonfUnoDos{} (\RCerEfectoUnoDos) & \RMontosWPBonfUnoDos{} (\RMontosEfectoUnoDos)\\
\quad Pipecat vs GPT-Live, $p$ ($r$) & \RWerWPBonfUnoTres{} (\RWerEfectoUnoTres) & \RCerWPBonfUnoTres{} (\RCerEfectoUnoTres) & \RMontosWPBonfUnoTres{} (\RMontosEfectoUnoTres)\\
\quad LiveKit vs GPT-Live, $p$ ($r$) & \RWerWPBonfDosTres{} (\RWerEfectoDosTres) & \RCerWPBonfDosTres{} (\RCerEfectoDosTres) & \RMontosWPBonfDosTres{} (\RMontosEfectoDosTres)\\
\botrule
\end{tabular}
\footnotetext[1]{Only the calls whose reference contains at least one amount: scenarios 1 and 4 have none.}
\footnotetext{Size of the reference: \RWerNA{}, \RWerNB{} and \RWerNC{} words, and
\RMontosNA{}, \RMontosNB{} and \RMontosNC{} amounts, in Pipecat, LiveKit and GPT-Live.
Tests: Kruskal--Wallis among the three architectures; Wilcoxon signed-rank test for
Pipecat versus LiveKit (paired by scenario and repetition) and rank-sum test for each
cascade against GPT-Live, with Bonferroni correction over the three pairs
($p_{\text{corr}}=\min(1,3p)$). Effect sizes: $\varepsilon^2_R=H/(N-1)$ for
Kruskal--Wallis and $r=Z/\sqrt{N}$ for each pair, with $N$ the calls compared (the
pairs, in the paired test)~\cite{peres2026}; an $r$ of 0.1, 0.3 and 0.5 is
conventionally read as small, medium and large.}
\end{table}

GPT-Live made fewer recognition errors (Table~\ref{tab:stt}): Pipecat made \RWerVecesA{}
times as many errors per word and LiveKit \RWerVecesB{} times as many. Kruskal--Wallis on
the WER of each call gives \RWerKwPTxt; with Wilcoxon and Bonferroni, Pipecat versus GPT-Live
gives \RWerWPBonfUnoTresTxt{} ($r=\RWerEfectoUnoTres$), LiveKit versus GPT-Live \RWerWPBonfDosTresTxt{} ($r=\RWerEfectoDosTres$) and Pipecat versus LiveKit \RWerWPBonfUnoDosTxt{} ($r=\RWerEfectoUnoDos$); the character error rate gives the same picture (Table~\ref{tab:stt}). Pipecat and LiveKit use the same recognizer, but each framework
sends it its own defaults. Two of those differences do not explain the gap between
them: Pipecat's profanity filter did not mask any word, and removing filler words (``eh'',
``mmm'') lowers the WER by the same amount in both, so that Pipecat versus LiveKit still
gives $p=\RSensPBonfSm$. GPT-Live had the lowest WER in all five scenarios.

The error on amounts is where the difference matters most, because those are the words that cost money.
Compared call by call, both cascades misheard more amounts than GPT-Live (Pipecat \RMontosWPBonfUnoTresTxt, $r=\RMontosEfectoUnoTres$; LiveKit \RMontosWPBonfDosTresTxt, $r=\RMontosEfectoDosTres$), and the two cascades did not differ from each other (\RMontosWPBonfUnoDosTxt).
The effect sizes are medium rather than large because each call contains only a few amounts, so most calls had no error on amounts in any architecture.
The errors of Pipecat and LiveKit are concentrated in scenario~5, with long amounts and self-corrections, and in
scenario~3, with repeated high amounts. When the response shows what the agent
understood (Table~\ref{tab:comprension}), Pipecat and LiveKit understood 18,000 almost
always (\RCompRetractadoA{} of \RCompVisibleA{} in Pipecat and \RCompRetractadoB{} of
\RCompVisibleB{} in LiveKit), and GPT-Live understood 18,500 in all of them
(\RCompCorregidoC{} of \RCompVisibleC). The error comes from the transcript and not
from the language model: Deepgram lost the ``quinientos'' (five hundred) in most Pipecat and LiveKit calls, and in every response that understood 18,000 the transcript said
``dieciocho mil'' without it. The language model repeated what it read. This is a
single line of the script, with \RCompNA{} calls per architecture: enough to see the
difference, not to estimate its size precisely.

\begin{table}[htbp]
\caption{Which amount the agent understood when the carrier said \emph{``dieciocho
mil\ldots{} no, dieciocho mil quinientos''} (scenario~5, turn~4)}\label{tab:comprension}
\setlength{\tabcolsep}{4pt}
\begin{tabularx}{\linewidth}{@{}l>{\hsize=0.9\hsize}Y>{\hsize=1.2\hsize}Y>{\hsize=1.0\hsize}Y>{\hsize=0.95\hsize}Y>{\hsize=0.95\hsize}Y@{}}
\toprule
& \textbf{Responses} & \textbf{Transcript with ``quinientos''} & \textbf{The response shows the amount} & \textbf{Understood 18,500} & \textbf{Understood 18,000}\\
\midrule
Pipecat & \RCompNA & \RCompSttA & \RCompVisibleA & \RCompCorregidoA & \RCompRetractadoA\\
LiveKit & \RCompNB & \RCompSttB & \RCompVisibleB & \RCompCorregidoB & \RCompRetractadoB\\
GPT-Live & \RCompNC & \RCompSttC & \RCompVisibleC & \RCompCorregidoC & \RCompRetractadoC\\
\botrule
\end{tabularx}
\footnotetext{Transcript: the one received by the language model (in GPT-Live, the
service's transcript, which the delegate reads).}
\end{table}

\subsection{Turn-taking}\label{sec:res-turnos}

Each architecture fails in a different way. Pipecat's detector cut \RFalsosA{} times in
the middle of a line, almost one extra cut per line, but Pipecat joined the pieces
before answering, and there was only \REncimaA{} speech overlap in \RLlamadasA{}
calls. LiveKit cut in the middle of a line \RFalsosB{} times: its 0.55~s wait buys
fewer cuts at the price of waiting longer in every turn. GPT-Live does not allow its
cuts to be counted, and it was the only one in which the two voices overlapped with
some frequency: \REncimaC{} times, in \RLlamadasEncimaC{} of its \RLlamadasC{} calls.
In \RSolapesRetomoC{} of them the agent had paused in the middle of its response, the
harness took the 2~s of silence as the end of the turn and played the next line, and
the agent resumed its response just as the line began: it was the harness that talked
over the agent.

\subsection{Interruptions and backchannels}\label{sec:res-interr}

\begin{table}[htbp]
\caption{Interruption scenario: how often each agent stopped, how long it took, what it said when it resumed and which agreements it registered. Fractions are events in which the agent stopped out of the events played while it was speaking}\label{tab:interr}
\renewcommand{\arraystretch}{1.18}
\setlength{\tabcolsep}{5pt}
\begin{tabularx}{\linewidth}{@{}p{7.2cm}*{3}{>{\centering\arraybackslash}X}@{}}
\toprule
& \textbf{Pipecat} & \textbf{LiveKit} & \textbf{GPT-Live}\\
\midrule
Calls & \RIntLlamadasA & \RIntLlamadasB & \RIntLlamadasC\\
\midrule
\multicolumn{4}{@{}l}{\textbf{A. Stopping behaviour}}\\
\cmidrule(r){1-1}
\textbf{Interruption} (expected: stop and answer) & & & \\
\quad Stopped & \RIntIntFracA & \RIntIntFracB & \RIntIntFracC\\
\quad Time to stop, median (ms) & \RIntDetMedA & \RIntDetMedB & \RIntDetMedC\\
\quad\quad interquartile range & {\small\RIntDetQaA--\RIntDetQbA} & {\small\RIntDetQaB--\RIntDetQbB} & {\small\RIntDetQaC--\RIntDetQbC}\\
\addlinespace
\textbf{Backchannel} (expected: keep talking) & & & \\
\quad Stopped & \RIntApoFracA & \RIntApoFracB & \RIntApoFracC\\
\quad Spoke again within 8~s & \RIntApoRetomaA & \RIntApoRetomaB & \RIntApoRetomaC\\
\quad Pause length, median (ms) & \RIntApoPauMedA & \RIntApoPauMedB & \RIntApoPauMedC\\
\quad\quad interquartile range & {\small\RIntApoPauQaA--\RIntApoPauQbA} & {\small\RIntApoPauQaB--\RIntApoPauQbB} & {\small\RIntApoPauQaC--\RIntApoPauQbC}\\
\addlinespace
\textbf{Control} (nobody speaks: base rate) & & & \\
\quad Stopped on its own & \RIntCtlFracA & \RIntCtlFracB & \RIntCtlFracC\\
\midrule
\multicolumn{4}{@{}l}{\textbf{B. Resumptions and agreements}}\\
\cmidrule(r){1-1}
\textbf{Resumptions after a backchannel} & \RIntRetTotalA & \RIntRetTotalB & \RIntRetTotalC\\
\quad continued the sentence & \RIntRetCA & \RIntRetCB & \RIntRetCC\\
\quad repeated or restarted & \RIntRetRA & \RIntRetRB & \RIntRetRC\\
\quad said something new & \RIntRetNA & \RIntRetNB & \RIntRetNC\\
\quad took it as an answer & \RIntRetTA & \RIntRetTB & \RIntRetTC\\
\addlinespace
\textbf{Calls closed with an agreement} & \RIntAcuRegistradosA & \RIntAcuRegistradosB & \RIntAcuRegistradosC\\
\quad \textbf{without the carrier's acceptance} & \textbf{\RIntAcuNoAceptadosA} & \textbf{\RIntAcuNoAceptadosB} & \textbf{\RIntAcuNoAceptadosC}\\
\quad\quad after an ``ok'' spoken over the agent & \RIntAcuApoyoA & \RIntAcuApoyoB & \RIntAcuApoyoC\\
\botrule
\end{tabularx}
\footnotetext{Fractions count the events that played while the agent was speaking,
excluding those during the farewell. ``Spoke again'' counts, among the times the agent
stopped, those in which it spoke again within 8~s. LiveKit and GPT-Live times include the
path between the harness and the agent.}
\end{table}

Faced with an interruption, all three stopped and answered (Table~\ref{tab:interr}).
Pipecat stopped fastest (median of \RIntDetMedA~ms); the \RIntDetMedB~ms of LiveKit include
its transport, and GPT-Live took the longest (\RIntDetMedC~ms). Faced with a backchannel,
Pipecat and LiveKit almost always stopped: with their default settings they treated
the ``ajá'' as an interruption (\RIntApoFracA{} backchannels in Pipecat and
\RIntApoFracB{} in LiveKit) and spoke again after a median pause of
\RIntApoPauMedA~ms in Pipecat and \RIntApoPauMedB~ms in LiveKit. They never stopped in
a control: they deliver their response in full, without pauses. GPT-Live stopped at
\RIntApoFracC{} backchannels, but also in \RIntCtlFracC{} controls, with nobody
speaking: it makes pauses of about one second (median of \RIntCtlPauMedC~ms) in the
middle of its responses.

When resuming, GPT-Live continued the sentence where it had left off in \RIntRetCC{} of
\RIntRetTotalC{} resumptions, Pipecat in \RIntRetCA{} of \RIntRetTotalA{} and LiveKit in
\RIntRetCB{} of \RIntRetTotalB. The most common outcome in Pipecat was to take the
backchannel as the carrier's answer (\RIntRetTA{} cases), and GPT-Live did so in
\RIntRetTC. One example of each case, with the backchannel in brackets: Pipecat took it
as an answer in ``Carlos, le puedo ofrecer'' [Mjm] ``Entonces, para confirmar, le
ofrezco diecisiete mil quinientos pesos por el viaje de Monterrey a Laredo\ldots''
(``Carlos, I can offer you'' [Mm-hm] ``So, to confirm, I offer you seventeen thousand
five hundred pesos for the trip from Monterrey to Laredo\ldots''); GPT-Live continued in
``Me da gusto que lo vea así.'' [Ajá] ``Y para hacerlo todavía más atractivo, también
le puedo dar prioridad para la siguiente ruta disponible en la zona.'' (``I am glad you
see it that way.'' [Uh-huh] ``And to make it even more attractive, I can also give you
priority for the next available route in the area.''). The behaviour is what one would
expect from each architecture: in Pipecat and LiveKit the backchannel cuts the
response in progress and the language model writes a new one, with the ``ajá'' as one
more carrier turn; GPT-Live listens while it speaks and can resume where it was.

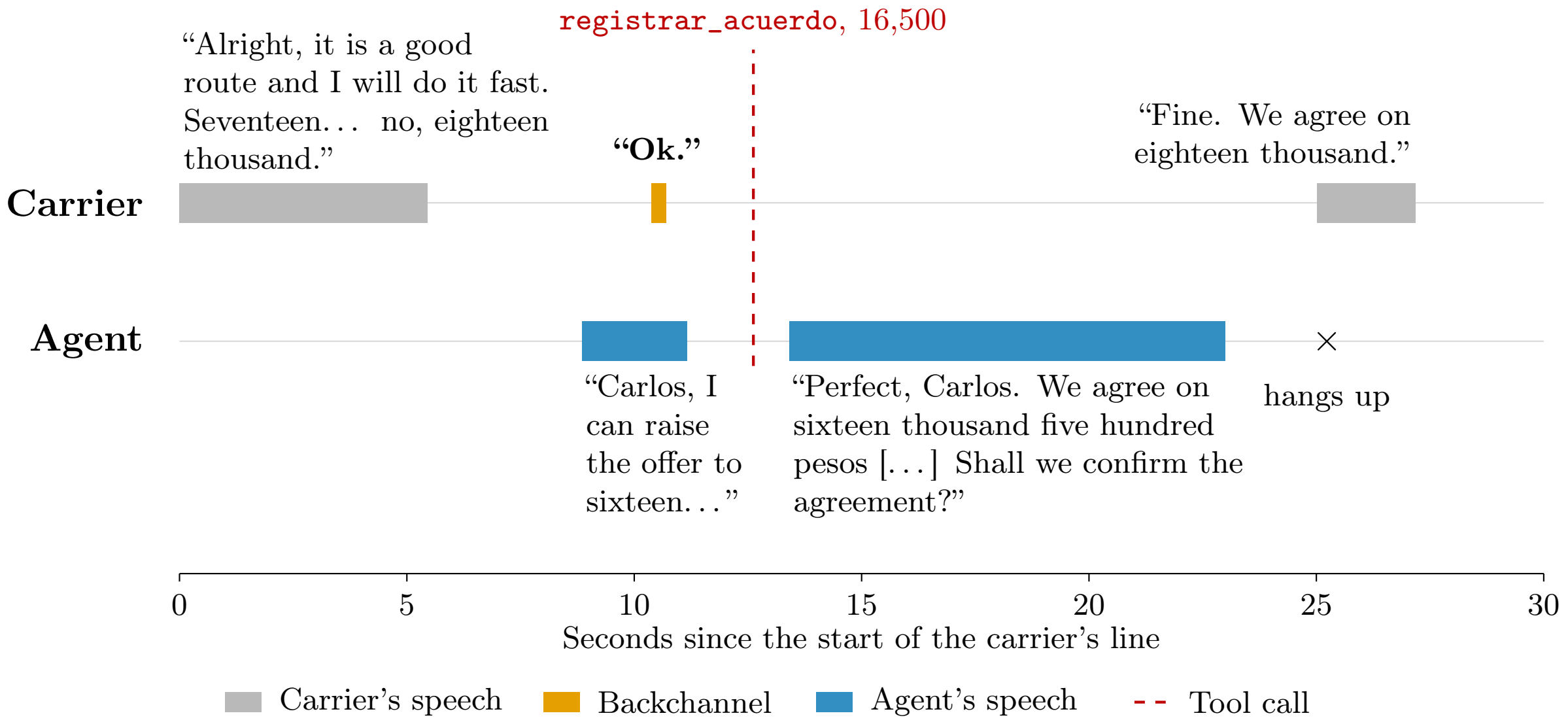
\begin{figure}[htbp]
\centering
\begin{fitwidth}
\begin{tikzpicture}[x=0.46cm, y=1cm, font=\footnotesize,
  dicho/.style={text width=#1, align=left, inner sep=1pt,
    execute at begin node={\hyphenpenalty=10000\exhyphenpenalty=10000}}]
\node[anchor=east, font=\normalsize\bfseries] at (-0.5,3.2) {Carrier};
\node[anchor=east, font=\normalsize\bfseries] at (-0.5,1.8) {Agent};
\draw[gray!30] (0,3.2) -- (\RLtEje,3.2);
\draw[gray!30] (0,1.8) -- (\RLtEje,1.8);
\fill[gray!55] (\RLtLineaIni,3.0) rectangle (\RLtLineaFin,3.4);
\fill[pocB] (\RLtApoyoIni,3.0) rectangle (\RLtApoyoFin,3.4);
\fill[gray!55] (\RLtSigIni,3.0) rectangle (\RLtSigFin,3.4);
\fill[pocA!80] (\RLtAntesIni,1.6) rectangle (\RLtAntesFin,2.0);
\fill[pocA!80] (\RLtDespuesIni,1.6) rectangle (\RLtDespuesFin,2.0);
\node at (\RLtCuelgaIni,1.8) {\normalsize$\times$};
\node[dicho=4.0cm, anchor=south west] at (\RLtLineaIni,3.5) {``Alright, it is a good route and I will do it fast. Seventeen\ldots{} no, eighteen thousand.''};
\node[anchor=south, font=\footnotesize\bfseries] at ({(\RLtApoyoIni+\RLtApoyoFin)/2},3.5) {``Ok.''};
\node[dicho=3.0cm, anchor=south east, align=right] at (\RLtSigFin,3.5) {``Fine. We agree on eighteen thousand.''};
\node[dicho=1.9cm, anchor=north west] at (\RLtAntesIni,1.5) {``Carlos, I can raise the offer to sixteen\ldots''};
\node[dicho=5.0cm, anchor=north west] at (\RLtDespuesIni,1.5) {``Perfect, Carlos. We agree on sixteen thousand five hundred pesos [\ldots] Shall we confirm the agreement?''};
\node[anchor=north] at (\RLtCuelgaIni,1.5) {hangs up};
\draw[red!75!black, dashed, thick] (\RLtRegistroIni,1.55) -- (\RLtRegistroIni,4.75);
\node[anchor=south, text=red!75!black] at (\RLtRegistroIni,4.75) {\texttt{registrar\_acuerdo}, \RLtMonto};
\draw (0,-0.55) -- (\RLtEje,-0.55);
\foreach \t in {0,5,...,\RLtEje} {\draw (\t,-0.55) -- (\t,-0.63); \node[anchor=north] at (\t,-0.63) {\t};}
\node[anchor=north] at ({\RLtEje/2},-0.95) {Seconds since the start of the carrier's line};
\begin{scope}[shift={(1,-1.85)}]
\fill[gray!55] (0,-0.1) rectangle (0.8,0.1); \node[anchor=west] at (0.9,0) {Carrier's speech};
\fill[pocB] (7,-0.1) rectangle (7.8,0.1); \node[anchor=west] at (7.9,0) {Backchannel};
\fill[pocA!80] (13,-0.1) rectangle (13.8,0.1); \node[anchor=west] at (13.9,0) {Agent's speech};
\draw[red!75!black, dashed, thick] (20,0) -- (20.8,0); \node[anchor=west] at (20.9,0) {Tool call};
\end{scope}
\end{tikzpicture}
\end{fitwidth}
\caption{The Pipecat call in which an ``ok'' spoken over the agent closed the deal (lines
translated from Spanish). Bars show each side's speech, coloured as in the legend.
The recognizer returned ``Ándele, una buena ruta y se la hago rápida. Diecisiete, no.''
and dropped the ``dieciocho mil'' (eighteen thousand). The agent stopped
\RLtCallaSeg~s after the ``Ok.'' began, registered the agreement \RLtRegistroSeg~s
later, before asking for confirmation, and hung up as the next line
started.}\label{fig:ok}
\end{figure}

Reviewing the resumptions revealed an effect that the scenario was not looking for: in
Pipecat and LiveKit, an ``ok'' or a ``sí, sí'' spoken over the agent could close the
deal. Pipecat registered the agreement right after a backchannel in \RIntAcuApoyoA{} of
its \RIntLlamadasA{} calls, LiveKit in \RIntAcuApoyoB{} and GPT-Live in none. In one Pipecat call
(Fig.~\ref{fig:ok}) the carrier had asked for 18,000; the agent began ``Carlos, puedo
subir la oferta a dieciséis\ldots'' (``Carlos, I can raise the offer to
sixteen\ldots''), an ``Ok'' was heard and the agent registered the agreement at
\RLtMonto. In the conversation log, the agent's message appears in full, with the
question ``¿Quedamos así, con ese monto y esas condiciones?'' (``Do we agree on that
amount and those conditions?''), even though only its beginning was heard, and the
``Ok.'' comes right after it as a carrier turn: for the language model, the backchannel
reads as the answer to that question. In total, an agreement was registered without
the carrier's acceptance in \RIntAcuNoAceptadosA{} of the \RIntAcuRegistradosA{}
agreements of Pipecat, \RIntAcuNoAceptadosB{} of the \RIntAcuRegistradosB{} of LiveKit and
\RIntAcuNoAceptadosC{} of the \RIntAcuRegistradosC{} of GPT-Live. These are counts over
\RIntLlamadasA{} calls per architecture and were not tested.

Of the eight hypotheses tested, only one passes the correction: \textbf{when it stops
at a backchannel, GPT-Live makes a shorter pause than LiveKit} (\RIntApoPauMedC{} versus
\RIntApoPauMedB~ms median, $p$~\RIntPruebaPausaTresDosP). That pause lasts as long as
the pauses it makes without a backchannel, so it does not show that the ``ajá'' causes
it. That GPT-Live stops less often than the other two, that the backchannel makes it stop
more often than the control ($p=\RIntPruebaApoyoControlTresP$) and that it continues
the sentence more often ($p=\RIntPruebaSigueTresUnoP$ against Pipecat and
$p=\RIntPruebaSigueTresDosP$ against LiveKit) point in the same direction, but these
results do not pass the correction. That Pipecat and LiveKit take the backchannel as an
answer more often than GPT-Live was not among the hypotheses and was not tested.

\section{Discussion}\label{sec:discusion}

Under identical, externally measured conditions, the three architectures differed less in latency than in how they handled spoken amounts and backchannels, and the latency that the frameworks reported about themselves hid most of the transport cost.
The paragraphs below take each part of this message in turn.

\paragraph{Why listening matters more than speed.} Three arguments support this reading of the results.
First, the consequences differ in kind: a misheard amount or an agreement closed on an ``ok'' changes the outcome of the negotiation, whereas the latency differences between the architectures were fractions of a second (Table~\ref{tab:diferencias}).
Second, a wait can be softened and a misheard word cannot: fillers make a delay feel shorter~\cite{shiwa2009,maslych2025,gonzales2025}, but no filler repairs an amount that the agent did not hear.
Third, the median wait of the three architectures, from \RObsMedC{} to \RObsMedB~ms, stayed below the 4~s beyond which delays degraded the quality of experience with LLM-powered virtual agents~\cite{maslych2025}, although in all three the 95th percentile exceeded that threshold (Table~\ref{tab:latencia}).
Listening while speaking therefore covers three abilities, and the architectures differed in all of them: hearing the words that matter, knowing when to stop and when to keep talking, and filling the wait while a tool runs.

\paragraph{Expected and observed latency.} GPT-Live was expected to be the
fastest, because it has no separate stages. Measured with the same instrument as the
other two, it cannot be said to be: with the median of all responses it is
\RDifAbsUnoTres~ms ahead of Pipecat, a difference that is
\verdict{\RDifVeredictoUnoTres}; with equal weight per line the two are
\verdict{\RLineaDifVeredictoUnoTres}; and without tools Pipecat is faster. GPT-Live's
overall median does not come from faster processing, but from using tools in fewer
responses and from the scenarios without amounts. What GPT-Live does do is make itself
heard sooner: it opens with a filler and responds to the first sound
\RPrimerDifUnoTresIc~ms before Pipecat. This difference between being heard and
answering could explain why a conversation with a native speech model feels more fluid
than its latency suggests, in line with earlier findings that fillers soften the perception of a delayed response~\cite{shiwa2009,maslych2025}.
In a recent study with embodied agents in virtual reality, conversational and gestural fillers during the wait improved the perceived response time and the perceived humanlikeness and naturalness of the agent~\cite{gonzales2025}; that study combined speech with gestures, whereas a telephone call carries only speech.
Nor is there a contradiction with the figure published for
GPT-Live: according to its own timestamps it took \RRelojModeloMed~ms, close to the
announced 0.8~s. The difference with what the carrier waited is the time the audio
takes to travel through the network and the OpenAI service and back, about
\RRelojServicioMed~ms and nearly constant, which those timestamps do not include: most
of it passes before the service returns the text of the last word of the line, and it
did not change when connecting over WebRTC (Section~\ref{sec:res-reloj}).

\paragraph{External measurement of latency.} The metrics that Pipecat and
LiveKit report about themselves fell short of the carrier's wait, by \RPropiaFaltaPctA\,\% in Pipecat and
\RPropiaFaltaPctB\,\% in LiveKit, and with them the two frameworks would have looked
the same. The figure published for GPT-Live falls short in the same way: it resembles
how long the model takes according to its own timestamps, not what the caller waits,
and the difference is the trip of the audio through the network and the service, the
largest part of that wait. To compare architectures, and to
monitor in production what the caller actually waits, latency has to be measured on
the listener's side and the instrument checked against a known delay, as was done
here. The protocol does not depend on the architecture: it is enough to connect the
harness to the new agent and repeat the validation over its audio path.

\paragraph{Composition of the wait.} Fig.~\ref{fig:capas} shows, stage by stage, where the time of a response goes, and three points stand out.
First, the tool call is the most expensive stage: in all three architectures it added \RRtDeltaMin{} to \RRtDeltaMax~s, more than any difference between frameworks.
Second, the stages that the frameworks orchestrate (end of turn, recognition, language model and synthesis) took similar median times in Pipecat and LiveKit, so what separates the two cascades is transport (\RTransporteA~ms in Pipecat against \RTransporteB~ms in LiveKit).
Third, most of GPT-Live's wait lies outside its own timestamps, in the trip of the audio through the network and the service.
A negotiation uses tools in many of its turns, so the changes with the greatest effect on the carrier's wait do not depend on the architecture: a faster LLM for the tools, validating prices in advance, or a brief filler as soon as a tool is called (``un segundo'', ``one second''), which would not shorten the wait but, according to the studies cited above, would make it feel shorter~\cite{maslych2025,gonzales2025}.
GPT-Live already starts to speak before it has the result in
\RHerrFrac\,\% of the responses that wait for one. That speed has a cost in control:
GPT-Live called the compensation tool \RFuncCompensaC{} times, against
\RFuncCompensaA{} and \RFuncCompensaB{} times for Pipecat and LiveKit, although it
offered compensations verbally as often as they did, and in \RAutCompSinHerrC{} of its
\RAutLlamadasC{} calls it stated at least one compensation that the tool did not record
(\RAutCompSinHerrA{} in Pipecat and \RAutCompSinHerrB{} in LiveKit); in one call it
offered \RAutSobreMaxMontoC{} pesos, above the maximum, after the tool had rejected the
amounts it tried. These counts were not tested, but they indicate that a model that
speaks while it waits for the tool can say what the tool did not validate.

\paragraph{Deployment and transport.} Compared with LiveKit, all of Pipecat's advantage
lies in transport, and part of it is the trip to a cloud server that Pipecat did not
make, because it ran on the same computer as the harness. In production all three
would also pay for the transport of the telephone network, so Pipecat's local
advantage would not carry over as it is. Much of LiveKit's latency depends on where
its media server is deployed, which the framework requires by design. A published
comparison of the two frameworks~\cite{chanl} describes this difference: LiveKit is
used with its cloud service or with a self-hosted server, whereas with Pipecat the
team provides the infrastructure and can develop and test everything on a single
computer, which is the condition under which Pipecat ran.

\paragraph{Recognition of spoken amounts.} Overall WER was low in
all three, but the error on amounts was not: Pipecat misheard \RMontosA\,\% of the
amounts and LiveKit \RMontosB\,\%, against \RMontosC\,\% for GPT-Live.
Table~\ref{tab:comprension} shows that the error reaches the agent's decisions: when the carrier corrected eighteen thousand to eighteen thousand five hundred, Pipecat went on with 18,000 in \RCompRetractadoA{} of the \RCompVisibleA{} responses that showed the amount and LiveKit in \RCompRetractadoB{} of \RCompVisibleB, whereas GPT-Live used 18,500 in \RCompCorregidoC{} of \RCompVisibleC.
In a negotiation a misheard amount costs money, and WER does not
show it, which agrees with earlier evidence that WER is a weak proxy for understanding and for the recognition of critical words~\cite{wang2003wer,benjannet2015}. The result corresponds to the chosen recognizer and its configuration, not to
cascades in general. If a cascade is chosen, it needs a safeguard, such as having the
agent repeat the amount and ask for confirmation before it calls
\texttt{proponer\_precio}.

\paragraph{Handling of backchannels.} Telling a backchannel from an attempt to take the floor is a long-standing problem in conversational systems~\cite{skantze2021,ekstedt2022vap}, and the results show what happens when it is not solved. With default settings, if the
carrier says ``ajá'', ``sí'' or ``ok'' while the agent is speaking, Pipecat and
LiveKit cut the sentence, may carry on as if the backchannel were an answer, and may
even close a deal that was not accepted. Both frameworks have options to keep a
backchannel from cutting the agent (a minimum number of words before yielding the turn,
or a mode that recognizes backchannels with a model); they were not tested, because the
study compares the default configuration, and a minimum number of words would also let
a flat ``no'' through. GPT-Live performed better without any configuration: it stopped
at fewer backchannels and, when it did stop, it continued the sentence in \RIntRetCC{}
of \RIntRetTotalC{} resumptions; but if it failed there would be nothing to tune.
Whatever the architecture, \texttt{registrar\_acuerdo} should check in code, and not
only request in its description, that the final amount was heard in full and that the
carrier accepted it with more than a backchannel spoken over the agent.

\subsection{Limitations}\label{sec:limitaciones}

The study has six main limitations.
First, the carrier was simulated.
All lines were recorded by a single person, with a headset microphone, and processed to sound like a call.
A real carrier speaks differently, interrupts and calls from noisy places, so the recognition results are an optimistic estimate.
The audio also did not go through the telephone network, which adds delay and packet loss.

Second, GPT-Live ran on another day, 18 days after Pipecat and LiveKit, because its campaign was repeated on the public API when the alpha version was withdrawn (Section~\ref{sec:poblacion}).
Its differences with Pipecat and LiveKit therefore include any change in the network or in the providers between those dates, which cannot be separated.
In addition, the GPT-Live figures correspond to the version of the model in service in October 2026 (\texttt{gpt-live-1}); another version could give different results.

Third, not all architectures heard the same lines.
Because of the branches and the point at which each agent closes the call, each architecture heard a different mix.
Between Pipecat and GPT-Live this matters: GPT-Live is ahead in the overall median (verdict: \verdict{\RDifVeredictoUnoTres}), whereas with equal weight per line the two are \verdict{\RLineaDifVeredictoUnoTres}.

Fourth, the harness did not hear GPT-Live as a telephone would.
The recording places each block of the GPT-Live voice at the time it arrived, whereas a telephone plays the blocks in sequence.
When that player is simulated with the bridge logs, the GPT-Live median goes from \RObsMedC{} to \RColaFilaMedC~ms, and the verdict for Pipecat~$-$~GPT-Live does not change (\RColaFilaDifUnoTres~ms: \verdict{\RColaFilaDifVeredictoUnoTres}).

Fifth, a single rater, J.~P.~M., classified the resumptions and the amount each agent understood.
How far two raters would agree is unknown, and the comprehension measure rests on one line of the script, with \RCompNA{} calls per architecture.

Sixth, each framework ran with a single configuration, its default one, and only in Spanish.
Other settings could change the turn-taking and backchannel results, and no script tested the switch to English that the instructions allow.

\section{Conclusion}\label{sec:conclusion}

This study compared Pipecat, LiveKit Agents and GPT-Live under the same conditions, with the same recorded carrier and an external harness, validated against a known delay, that measures what the listener waits.
The architectures differed most in how they listened.
GPT-Live misheard \RMontosC\,\% of the amounts, against \RMontosA\,\% in Pipecat and \RMontosB\,\% in LiveKit, and when the carrier corrected eighteen thousand to eighteen thousand five hundred, it used the corrected amount in every response that showed one, whereas the cascades mostly kept eighteen thousand.
With default settings, Pipecat and LiveKit stopped at backchannels and often took them as an answer, to the point of registering agreements that the carrier had not accepted, whereas all three yielded the floor to a real interruption.
Latency separated them less.
No architecture was clearly faster for the content response: GPT-Live was \RDifAbsUnoTres~ms ahead of Pipecat in the overall median (95\,\% CI from \RDifLoUnoTres{} to \RDifHiUnoTres~ms), LiveKit's extra wait lay in its transport, and a tool call cost about one second in all three architectures, more than any difference between frameworks.
The latency that Pipecat and LiveKit report about themselves fell short of the listener's wait by \RPropiaFaltaPctA\,\% and \RPropiaFaltaPctB\,\%, and the 0.8~s published for GPT-Live matches what its own timestamps measure (\RRelojModeloMed~ms), not the \RObsMedC~ms that the listener waited, so latency must be measured from outside the agent.
These results hold for one recorded speaker, audio without the telephone network and the default configuration of each framework; cost, platform maturity and integration with SPOTS were outside the scope of this study.

Future work will repeat the measurements with real telephone calls and with several voices, accents and noise levels, and will test the changes that do not depend on the architecture: a faster LLM for the tools, validating prices in advance and a brief filler while a tool runs.
For the cascades, it will add a confirmation of every amount before \texttt{proponer\_precio} is called, measure how much this reduces the error, and test the Pipecat and LiveKit options that keep a backchannel from cutting the agent; in every architecture, \texttt{registrar\_acuerdo} will check in code that the amount was heard in full and accepted.
Finally, a second rater will classify the resumptions and the understood amounts without seeing the first rating, and the provider will be asked in which part of its service the \RRelojServicioMed~ms that GPT-Live's timestamps do not include are spent.

\backmatter

\bmhead{Supplementary information}

The Supplementary Material contains the agent's instructions, the trip data and the
tools (S1), all carrier lines by scenario, with the branches and the lines of the
interruption scenario (S2), and the eight hypotheses tested in the interruption scenario, with their $p$-values (S3).

\section*{Declarations}
\label{sec:declarations}

\textbf{Funding.} This research was conducted as part of the authors' work in the AI Lab at Freight Technologies Inc. and received no external grant funding.

\textbf{Competing interests.} Jovan Popoca Moran, Eduardo de Avila-Armenta, Ra\'ul Alejandro Aquino-Ibarra, Jorge Octavio Nicol\'as-D\'iaz, and Umberto Leon-Dominguez are employed by Freight Technologies Inc., a company that develops automated freight-brokering and negotiation technology directly related to the subject matter of this paper. This relationship is disclosed as a potential competing interest.

\textbf{Data and code availability.} The code and data supporting this study are openly available at \url{https://github.com/AILabFr8tech/voice-negotiation-agents} under the Apache License 2.0. The repository contains the three agent architectures (Pipecat, LiveKit and GPT-Live), the shared negotiation engine, system prompt and tools, the harness that played the carrier's recorded lines, the simulated agents used to validate the latency and interruption instruments, and the analysis code. It also contains the exported per-call and per-turn tables from which every result is computed, and the scripts that regenerate each number and figure in this paper from those tables; the numbers in the text are written by those scripts, not by hand. The audio recordings of the carrier's lines and the raw per-call logs and transcripts are not included, because they contain the voice of the person who recorded the lines; the repository lists the cryptographic hash of every audio clip, which shows that the three architectures heard the same audio. API keys for the commercial services are not included, and running the agents against those services requires the user's own accounts.



\end{document}